\documentclass{article}     

\usepackage{arxiv}
\usepackage[utf8]{inputenc} 
\usepackage[T1]{fontenc}    
\usepackage{hyperref}       
\usepackage{url}            
\usepackage{booktabs}       
\usepackage{amsfonts}       
\usepackage{nicefrac}       
\usepackage{microtype}      
\usepackage{lipsum}
\usepackage{graphicx}
\usepackage{caption}
\usepackage{subcaption}
\usepackage{multirow}
\usepackage{caption}
\usepackage{array}
\usepackage{float}
\usepackage{textcomp}
\usepackage{amsmath}
\usepackage{amssymb}
\usepackage{stackrel}
\usepackage{algorithmic}
\usepackage{systeme}
\usepackage{color}
\usepackage{enumerate}
\usepackage{bbm}
\usepackage{wrapfig}
\DeclareMathOperator{\atan}{atan}
\DeclareMathOperator{\atantwo}{atan2}
\title{Overconstrained Locomotion}
\author{
  Haoran Sun\\
  Department of Mechanical and Energy Engineering, Southern University of Science and Technology\\
  Department of Computer Science, The University of Hong Kong\\
  \And
  Bangchao Huang, Zishang Zhang, Ronghan Xu, Guojing Huang, Shihao Feng, Guangyi Huang\\
  Department of Mechanical and Energy Engineering, Southern University of Science and Technology\\
  \And
  Jiayi Yin, Nuofan Qiu\\
  School of Design\\
  Southern University of Science and Technology\\
  \And
  Hua Chen\\
  ZJU-University of Illinois Urbana-Champaign Institute\\
  Zhejiang University\\
  \And
  Wei Zhang\\
  System Design and Intelligent Manufacturing\\
  Southern University of Science and Technology\\
  \And
  Jia Pan\\
  Department of Computer Science\\
  The University of Hong Kong\\
  \And
  Fang Wan$^{*}$\\
  School of Design\\
  Southern University of Science and Technology\\
  \texttt{wanf@sustech.edu.cn}\\
  \And
  Chaoyang Song\thanks{Corresponding Authors.}\\
  Design \& Learning Research Group\\
  Southern University of Science and Technology\\
  \texttt{songcy@ieee.org}\\
}
\begin{document}
\maketitle
\begin{abstract}

    This paper studies the design, control, and learning of a novel robotic limb that produces overconstrained locomotion by employing the Bennett linkage for motion generation, capable of parametric reconfiguration between a reptile- and mammal-inspired morphology within a single quadruped. In contrast to the prevailing focus on planar linkages, this research delves into adopting overconstrained linkages as the limb mechanism. The overconstrained linkages have solid theoretical foundations in advanced kinematics but are under-explored in robotic applications. This study showcases the morphological superiority of Overconstrained Robotic Limbs (ORLs) that can transform into planar or spherical limbs, exemplified using the simplest case of a Bennett linkage as an ORL. We apply Model Predictive Control (MPC) to simulate a range of overconstrained locomotion tasks, revealing its superiority in energy efficiency against planar limbs when considering foothold distances and speeds. The results are further verified in overconstrained locomotion policies optimized from Reinforcement Learning (RL). From an evolutionary biology perspective, these findings highlight the mechanism distinctions in limb design between reptiles and mammals and represent the first documented instance of ORLs outperforming planar limb designs in dynamic locomotion. Future studies will focus on deploying the model-based and learning-based overconstrained locomotion skills in the robotic hardware to close the Sim2Real gap for developing evolutionary-inspired, energy-efficient control of novel robotic limbs. 
    
\end{abstract}
\section{Introduction}
\label{sec:Introduction}

    Designing the limb mechanism has been widely understood as setting the legged robot's upper boundary capability in efficient locomotion, where the current literature has shown a maturing adoption of the four-bar linkage or its variations for a practical solution~\cite{Biswal2021Development}. Theoretical kinematics suggests other solutions, such as the overconstrained or the spherical designs, that also share a minimum of four revolute joints for a mobile linkage besides the planar case~\cite{Gu2023Computational}. Biological inspiration from nature suggests a collection of versatile limb mechanisms concerning the body designs in omnidirectional locomotion on challenging terrains~\cite{Iida2016Biologically}. There remains a research gap to identify an optimal design of robotic limbs for omnidirectional and energy-efficient locomotion where a comparative study is difficult to implement in practice due to the challenges of achieving different limb morphologies within a single robotic system.

    \begin{figure}[htbp]
        \centering
        \includegraphics[width=0.65\linewidth]{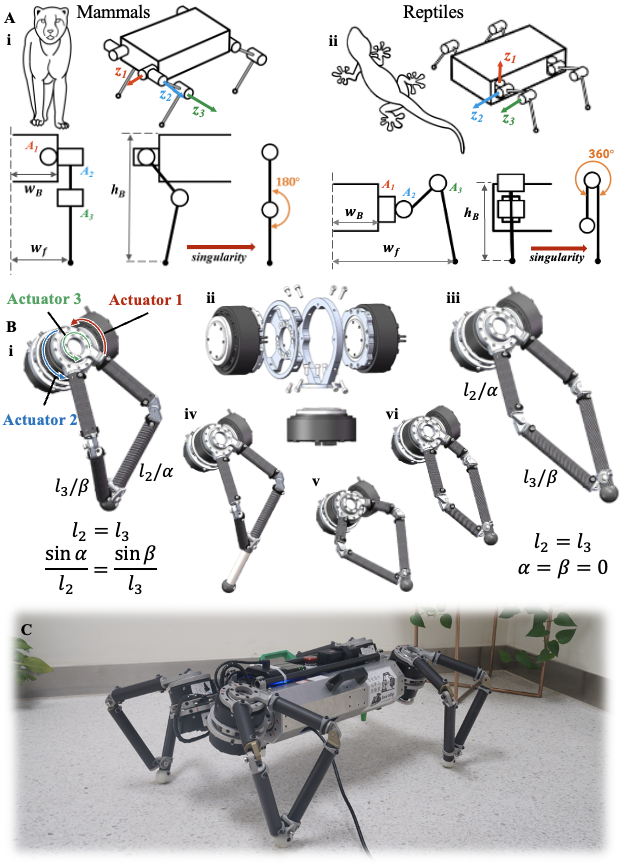}
        \caption{
        \textbf{Morphological inspiration from mammal and reptile limbs to the modern design of quadrupedal robotic limbs using closed-loop linkages.}
        (A) Schematic illustrations of quadruped limb designs inspired by i) mammals and ii) reptiles, where $z_i (i=1,2,3)$ is the axis of actuator $A_i$, $w_B$ is the body width, $w_f$ is the foot-ground contact width, and $h_B$ is the robot height. Both limb designs usually have a kinematic singularity, typical for the limbs to stretch straight for mammals and be folded for reptiles.
        (B) Parametric design reconfiguration of the robotic limbs, including i) the Overconstrained Robotic Limb with a Bennett linkage, ii) the triple-QDD module, iii) the planar 4-bar robotic limb with equal length in opposite links, and design variants of the robotic limb with iv) extended foot, v) diamond-shaped planar 4-bar, and vi) slender-parallelogram planar 4-bar.
        (C) The engineering prototype in standing posture.
        }
        \label{fig:Intro_PaperOverview}
    \end{figure}

    The modern design of robotic limbs usually draws morphological inspirations from animals, such as mammals and reptiles. As shown in Fig.~\ref{fig:Intro_PaperOverview}, most quadrupedal mammals feature a relatively higher center of gravity (COG) for agility with limbs stretching out to the ground directly from the body~\cite{Ijspeert2020Amphibious}, resulting in a comparable ratio between the ground contact width $w_f$ between the thoracic or pelvic limbs against the body width $w_B$. Reptile animals usually feature a lower COG for stability~\cite{Kitano2016Titan} and reduced power consumption~\cite{Tsunoda2022Power} with the limbs spanning to the side of the body, resulting in a much larger ratio between $w_f$ and $w_B$. They both leverage the kinematic singularity for more energy-efficient support of the body weight during locomotion, with the mammal limbs capable of stretching to almost 180 degrees and the reptile limbs foldable to nearly 360 degrees. 
    
\begin{figure}[htbp]
        \centering
        \includegraphics[width=0.8\linewidth]{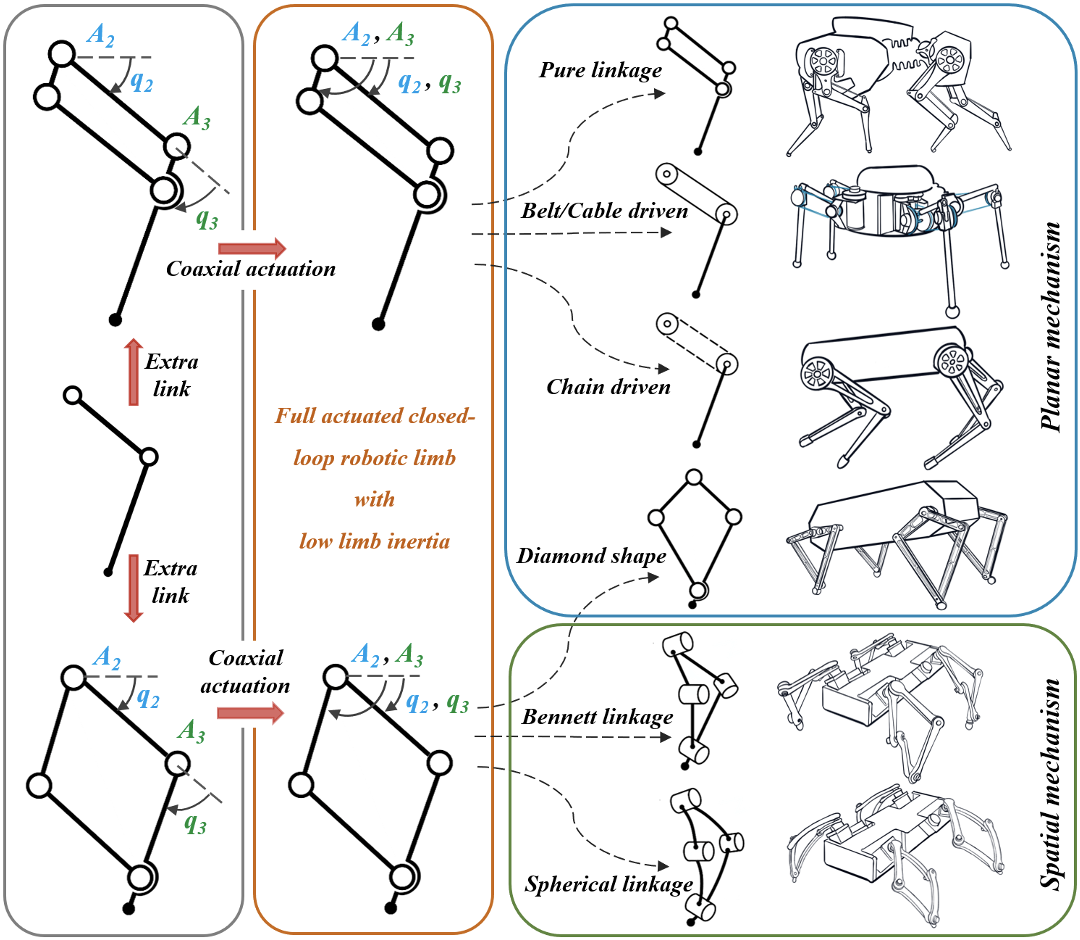}
        \caption{
        \textbf{Generalization of the co-axial, dual-actuator design of the robotic limbs.} Here, actuators $A_2$ and $A_3$ drive a planar four-bar linkage or its variations for a more compact center of gravity, which is reflected among a wide range of existing quadrupedal robots with engineering possibilities for new limb designs if spatial four-bar linkages are adopted.
        }
        \label{fig:Intro_LitRev}
    \end{figure}

    Such biological features are widely adopted in the modern design of robotic limbs, where a co-axial, dual-output actuation using closed-loop linkages or their variations~\cite{Kalouche2016Design, Seok2014Design, Hubicki2016ATRIAS} are widely used instead of a direct open-loop design~\cite{Hutter2016Anymal}, as shown in Fig.~\ref{fig:Intro_LitRev}. The parallelogram mechanism is a typical choice in robotics, offering kinematic attributes and energy efficiency similar to a serial chain when one link attached to the actuator is significantly shorter than the other~\cite{Hooks2020Alphred}. Additionally, robotic limbs driven by belts~\cite{Kitano2016Titan} or chains~\cite{Bledt2018Cheetah} have been proposed, which are kinematically equivalent to a planar four-bar linkage as a robotic limb. Some quadrupeds, such as Ghost Robotics Minitaur~\cite{Daniel2016Gait} and Stanford Doggo~\cite{Kau2019Stanford}, employ planar 4\textit{R} linkages of diamond shapes without using actuator $A_1$ on the hip, sacrificing the efficiency in omnidirectional locomotion but excel in an elegant and straightforward design of robotic limbs at a much-reduced cost of integration. Recent work also shows the design possibility of using spatial four-bar linkages such as the overconstrained Bennett linkage~\cite{Carvalho2016MotionAnalysis, Feng2021Overconstrained, Gu2022Overconstrained} or spherical linkage~\cite{Gu2023Computational} with enhanced energy efficiency in omnidirectional locomotions. 

    This paper investigates the overconstrained design of quadrupedal robotic limbs driven by quasi-direct drives (QDD) for comparative analysis regarding its energy efficiency in omnidirectional locomotion against the planar limbs. While the body frames and actuators adopt a similar architecture to quadrupeds driven by QDDs, we design the limbs to be reconfigurable between an overconstrained linkage and a planar one within a single modular system. We conduct kinematic and dynamic modeling of the limbs and validate their locomotion performance by employing model predictive control in simulations. We confirm the overconstrained robotic limbs' quadrupedal locomotion capabilities across various tasks, including forward, lateral, and turning movements, traversal on challenging terrains like gravel, slopes, and stairs, and their ability to recover from lateral disturbances. We also implement reinforcement learning to train overconstrained locomotion in flat and multi-terrain environments. By resetting all twist angles to zero, using precisely the same robotic system, we facilitate a direct comparison with conventional limb designs featuring planar four-bar linkages, further highlighting the superior energy-efficient, omnidirectional quadrupedal locomotion achieved by our overconstrained robotic limb design. This work offers valuable insights into mechanism design for legged robots employing overconstrained spatial linkages, the contributions of which are listed below.
    \begin{enumerate}
        \item Proposed the novel design of an overconstrained robotic limb actuated by Quasi-Direct Drives that achieves high energy efficiency in omnidirectional locomotion while enabling morphology transformation.
        \item Developed a Model Predictive Control framework for overconstrained locomotion using the reconfigurable Bennett limb and validated its favorable motion performance and terrain adaptability through simulation.
        \item Trained overconstrained locomotion policies with Reinforcement Learning using the reconfigurable Bennett limbs and verified its superior energy efficiency over planar limbs. 
    \end{enumerate}

    The rest of this paper is organized as follows. Section~\ref{sec:Design} presents the QDD design of the overconstrained robotic limb (ORL) parametrically reconfigurable as a mammal- or reptile-inspired quadruped. Section~\ref{sec:Limb} presents the motion control and experimental validation of a single overconstrained robotic limb. Simulated results for overconstrained locomotion are implemented using Model Predictive Control in Section~\ref{sec:MPC} with comparative energy efficiency analysis and reinforcement learning in Section~\ref{sec:RL}. The conclusion, limitations, and future work are in the final section. 

\section{Overconstrained Robotic Limb Design}
\label{sec:Design}

\subsection{Kinematic Analysis}

\begin{figure}[htbp]
        \centering
        \includegraphics[width=0.55\linewidth]{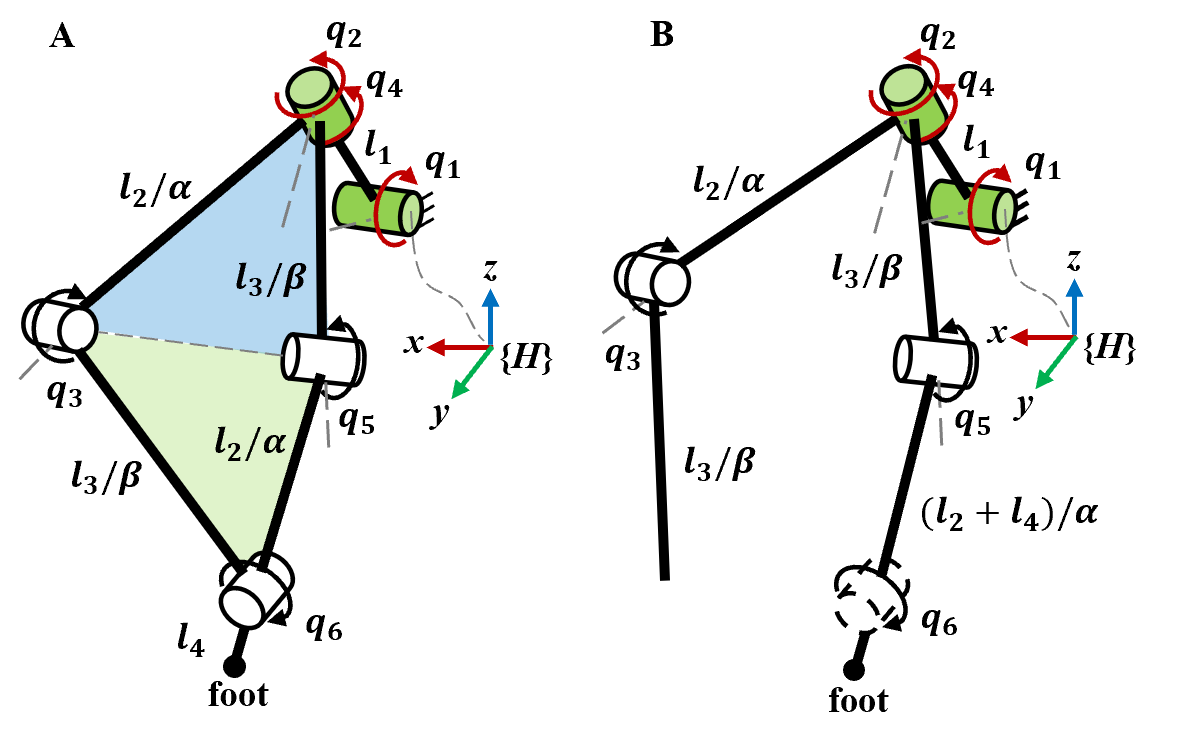}
        \caption{
        \textbf{The schematic diagrams of overconstrained robotic limbs with the green and white cylinders as active and passive joints, respectively.}
        (A) The closed-loop model. 
        (B) The spanning tree model.}
        \label{fig:Design_Kinematics}
    \end{figure}

    Shown in Fig.~\ref{fig:Design_Kinematics}A is the kinematic diagram of an ORL. The first joint is grounded to the robot chassis, providing hip motion \(q_1\) actuated by a QDD, which is linked to two co-axial dual-output QDD joint couple \(q_2, q_4\) through link \(l_1\). A Bennett linkage is attached to the QDD joint couple with link \(l_2\) actuated by both joints \((q_2, q_4)\) and link \(l_3\) actuated by joint \(q_2\). The Bennett linkage passively actuates the rest of the joints \((q_3, q_5, q_6)\). We denote the twist angles as \(\alpha\) for the link of length \(l_2\) and \(\beta\) for the link of length \(l_3\), which are constrained by the overconstrained geometric condition of a Bennett linkage as \(\sin{\alpha}/l_2 = \sin{\beta}/l_3\), or the Bennett Ratio. In addition, we extend the link between \(q_5\) and \(q_6\) with length \(l_4\) to represent the foot at the end. 
    
    The spanning tree model shown in Fig.~\ref{fig:Design_Kinematics}B shares a closer resemblance to a robotic limb with a serial chain, which is kinematically equivalent to the closed-loop model by considering joint \(q_5\) as the knee joint, virtually actuated by the Bennett linkage from the joint couple \(q_2, q_4\). The portion above the knee is the thigh denoted kinematically as \(l_3/\beta\) in Fig.~\ref{fig:Design_Kinematics}B, and the portion below the knee is the leg-foot denoted kinematically as \((l_2+l_4)/\alpha\). 

    We computationally optimize the link lengths $l_2$ and $l_3$ for maximum workspace under the constraint of a total leg length using the method similar to ~\cite{Feng2021Overconstrained}, resulting in a design of $l_2 = l_3 = 0.18m$ based on the engineering specifications listed in the Supplementary Materials. Considering the payload capability and the workspace, the twist angles are selected as $\alpha = 120^{\circ}$ and $\beta = 60^{\circ}$, respectively. Under the given geometry conditions, the explicit closure equations~\cite{Baker1979theBennett} of the single-leg mechanism are listed below,
    \begin{align}
        q_3 - q_5 &= 0, \label{eq:closure_eq1} \\
        \tan{(\frac{\pi - (q_4 - q_2)}{2})} \tan{(\frac{q_3}{2})} &= \csc{(\frac{\beta - \alpha}{2})}, \label{eq:closure_eq2}
    \end{align}
    which has six joints in total, including three actuated joints ($q_1, q_2, q_4$) and three passive ones ($q_3, q_5, q_6$). To simplify the model, we use the spanning tree model in Fig.~\ref{fig:Design_Kinematics}B to facilitate the analysis as an equivalent serial limb. As a result, we can model the ORL by mapping the motor space to the joint space of the spanning tree model and then mapping the spanning tree's joint space to the foot's Cartesian space.

\subsection{Engineering Design}
    
\begin{figure}[htbp]
        \centering
        \includegraphics[width=0.6\linewidth]{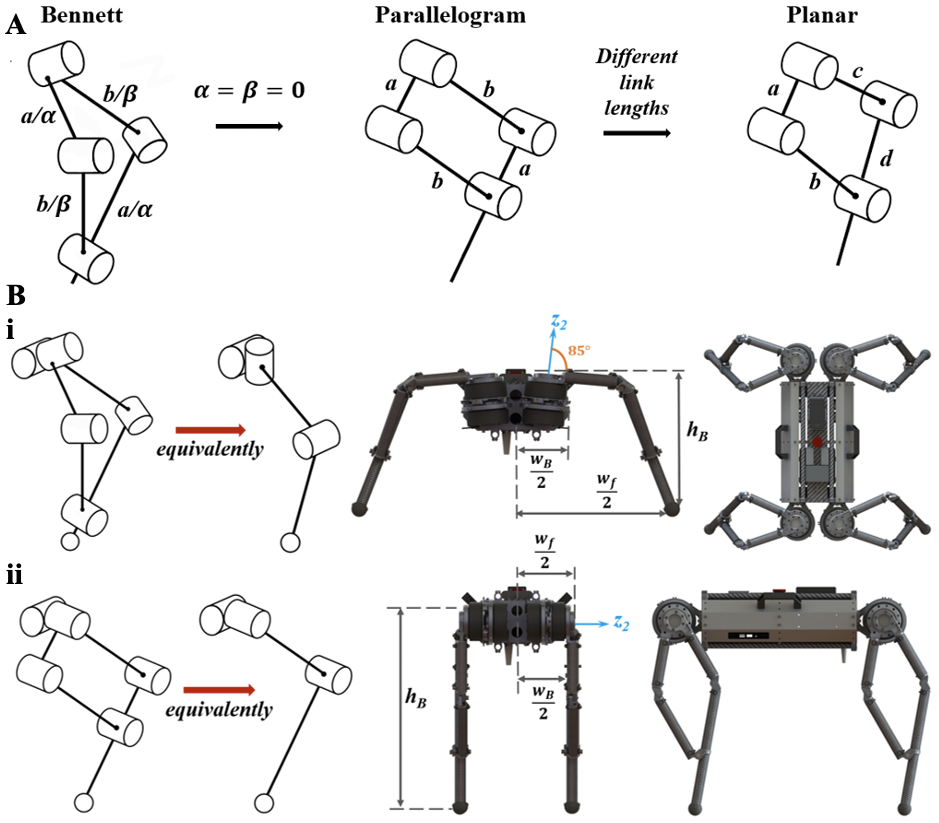}
        \caption{
        \textbf{Parametric reconfiguration of the overconstrained robotic limb.}
        (A) Parametric reconfiguration from a Bennett linkage to a parallelogram by constraining all twist angles to zeros, then to a planar limb with different link lengths, where the extension for foot length could be added at any time to enable morphological reconfiguration of the overconstrained quadruped. 
        (B) Exemplified illustration of the engineering prototype parametrically reconfigured into i) reptile- or ii) mammal-inspired morphology.
        }
        \label{fig:Design_Reconfig}
    \end{figure}

    The resultant overconstrained quadruped is shown in Fig.~\ref{fig:Intro_PaperOverview}C. As shown in Fig.~\ref{fig:Intro_PaperOverview}B, its actuator arrangement is consistent with a conventional mammal-type quadruped but replaced with an overconstrained Bennett linkage as the limb, attached to a unified chassis for comparative analysis, resulting in a coupled fully three-dimensional foot force generation. The three QDD actuators drive the ORL with a co-axial, dual-output actuation, resulting in a coupled limb movement, which can be described as follows,
    \begin{equation}
        \begin{bmatrix}
            \tau_1 \\
            \tau_2 \\
            \tau_3
        \end{bmatrix} = \begin{bmatrix}
            1 & 0 & 0 \\
            0 & 1 & -1 \\
            0 & 0 & 1
        \end{bmatrix} \begin{bmatrix}
            \tau'_1 \\
            \tau'_2 \\
            \tau'_3
        \end{bmatrix},
        \label{eq:tau2tau_actuator}
    \end{equation}
    where $\tau'_1$, $\tau'_2$, and $\tau'_3$ are the actuator torque from the actuators 1, 2, and 3 shown in Fig.~\ref{fig:Intro_PaperOverview}B, and $\tau_1$, $\tau_2$, and $\tau_3$ are the active joint torques related to $q_1$, $q_2$, and $q_3$ respectively in the kinematic model illustrated in Fig.~\ref{fig:Design_Kinematics}. As shown in Fig.~\ref{fig:Design_Reconfig}A, the carbon fiber link design of the ORL is parametrically reconfigurable into a reptile- or mammal-inspired design shown in Fig.~\ref{fig:Design_Reconfig}B. Please refer to the Supplementary Video for further elaboration.

\subsection{Bio-inspired Morphological Generalization}

    For comparative analysis, it is vital to establish a line of kinematic reconfiguration among the three types of four-bar linkages with revolute joints only, including the planar, overconstrained, and spherical cases, for a parametric generalization towards a reconfigurable link design. 
    
    For an overconstrained 4\textit{R} linkage to be parametrically reconfigured into a planar 4\textit{R} case, we can start by setting all twist angles to zeros to achieve a planar design immediately. We can further relax the requirement on the link length to make the resultant planar 4\textit{R} more general in representation. Similarly, for an overconstrained 4\textit{R} linkage to be parametrically reconfigured into a spherical 4\textit{R} case, we can start by setting all link lengths to zeros to achieve a spherical design immediately. Then, we can further relax the requirement on the twist angles to make the resultant spherical 4\textit{R} more general in representation (See the Supplementary Video for further elaboration). However, the reversed transformation from the planar or spherical cases to the overconstrained 4\textit{R} linkage requires adding constraints to multiple parameters simultaneously, which is not as straightforward as the other way around. Morphologically speaking, the overconstrained 4\textit{R} linkage (spatial representation using Cartesian coordinates) preserves critical geometric constraints that could be dimensionally reduced to the planar case using the 2D coordinates or the spherical case using the spherical coordinates. 
        
    Fig.~\ref{fig:Design_Reconfig}A demonstrates the parametric reconfiguration process using the proposed limb, starting with a Bennett linkage as the core design of an ORL with near zero foot length. By fixing all twist angles to zeros, i.e., \(\alpha = \beta = 0\), we get a planar design of the limb, which can be modified kinematically to add a longer foot length, resulting in an ORL of parallelogram design. It shares a closer resemblance to the limb design shown in~\cite{Hooks2020Alphred} when the link attached to actuator 2 is much shorter than the other link attached to the actuator 3, i.e., \(a < b\). One can make it a more generalized design of a planar case by changing the link lengths to take different values, which is viable in theory but less adopted in practice due to various engineering reasons. As a result, using a unified body and actuation system, we can quickly reconfigure the whole robot design to make it a versatile system with reconfigurable limbs as a reptile- or mammal-inspired quadruped in Fig.~\ref{fig:Design_Reconfig}B for comparative analysis, with or without the extended foot. 
    
\section{Motion Control for an Overconstrained Limb}
\label{sec:Limb}

\begin{figure}[htbp]
        \centering
        \includegraphics[width=1.0\linewidth]{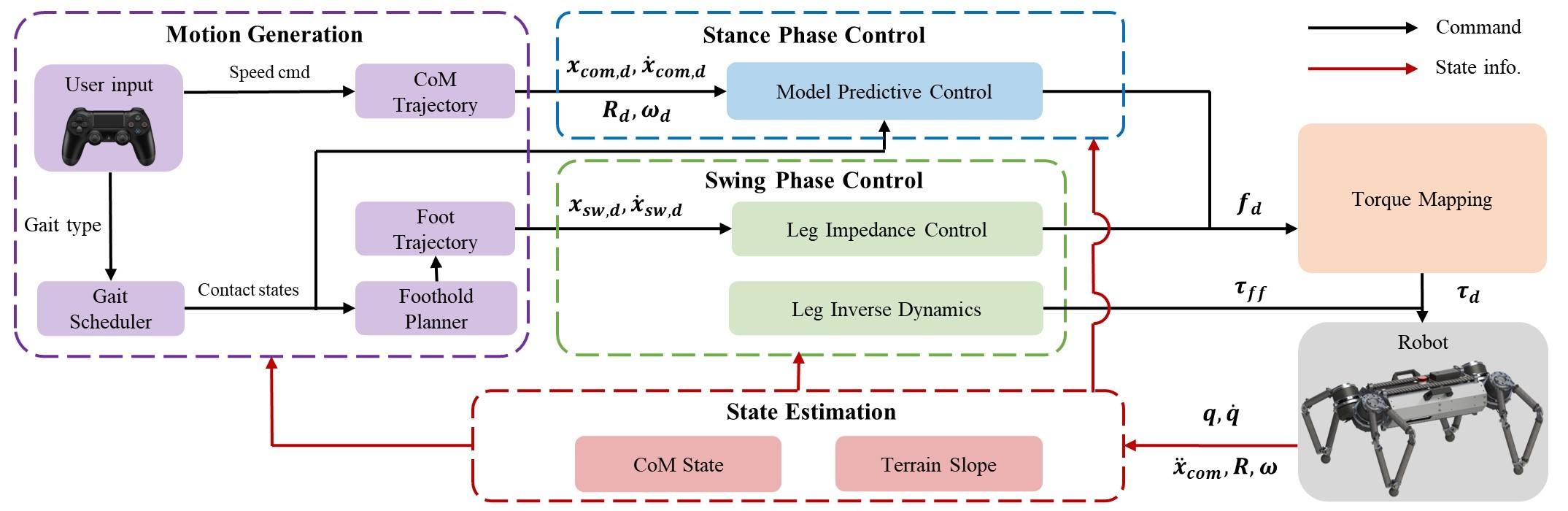}
        \caption{
        \textbf{The locomotion control framework of the overconstrained quadruped.}
        }
        \label{fig:Limb_MotionControl}
    \end{figure}

    The locomotion control framework of the overconstrained quadruped is illustrated in Fig.~\ref{fig:Limb_MotionControl}. The control framework mainly includes state estimation, motion generation, stance phase control, and swing phase control. Please refer to the appendix for more details.

\subsection{Leg Swing Control}
\label{subsec:swing_control}

    During the swing phase, the foot would liftoff and move to the desired foothold location that maintains the desired velocity and keeps the robot stable. The foothold locations of each leg are calculated from a neutral point utilizing a linear combination of the Raibert heuristic~\cite{Raibert1986Legged} and a velocity feedback term. The selection of a neutral point is usually related to the hip location. The neutral point of the overconstrained quadruped is chosen laterally outwards with the hip location to maintain an energy-efficient leg configuration. Thus, the foothold location on the flat plane of $j_{th}$ leg is calculated by
    \begin{equation}
        p_{end,j} = p_{h,j} + 
        \begin{matrix} \underbrace{
            \frac{v_d T_{st}}{2} + \left( R_z(\frac{\omega_d T_{st}}{2}) - \mathbb{I}_3 \right) p_{h,j} } \\ Raibert
        \end{matrix}
        + K_v (v - v_d).
    \end{equation}
    Considering the balanced GRF of the front and rear legs in slope terrain, the following adjustments of foothold location should be added 
    \begin{equation}
        \Delta p = \begin{bmatrix}
            h_d \tan{\theta_{d}} \\
            -h_d \tan{\phi_{d}}
        \end{bmatrix},
    \end{equation}
    where $p_{h,j} \in \mathbb{R}^2$ is the neutral point, $T_{st} \in \mathbb{R}$ is the scheduled stance time, $K_v \in \mathbb{R}^{2 \times 2}$ is the feedback gain which can also be replaced to capture point gain~\cite{Pratt2006Capture}, $h_d \in \mathbb{R}$ is the desired height, $\theta_{d} \in \mathbb{R}$ is the desired pitch, and $\phi_{d} \in \mathbb{R}$ is the desired roll. The calculations for the foothold are expressed in the body frame. 

    A cycloidal function generates the foot trajectory between the foothold and liftoff locations. The acceleration of the start and end points of the cycloid trajectory is zero, which can reduce the impact caused by sudden velocity changes at the foot. A Cartesian impedance controller with a feedforward term calculates the joint torque required by the foot trajectory tracking. The full controller is formulated as
    \begin{equation}
        \tau = \tau_{ff} + J^T [K_{p} (p_d - p) + K_{d} (\Dot{p}_d - \Dot{p})],
    \end{equation}
    where $J$ is the jacobian, $K_{p}$ and $K_{d}$ are the diagonal matrices of stiffness and damping gain, and $\tau_{ff}$ is the joint feedforward torque computed by the inverse dynamics of the ORL, which is derived in detail in the supplementary material.

\subsection{Experimental Validation for Leg Swing}    

    The foot's position and force tracking experiments are implemented to evaluate the ORL's performance. The prototype's control bandwidth is set to 250 Hz.

    \begin{wrapfigure}{r}{0.6\textwidth}
        \vspace{-1.7cm}
        \centering
        \includegraphics[width=0.9\linewidth]{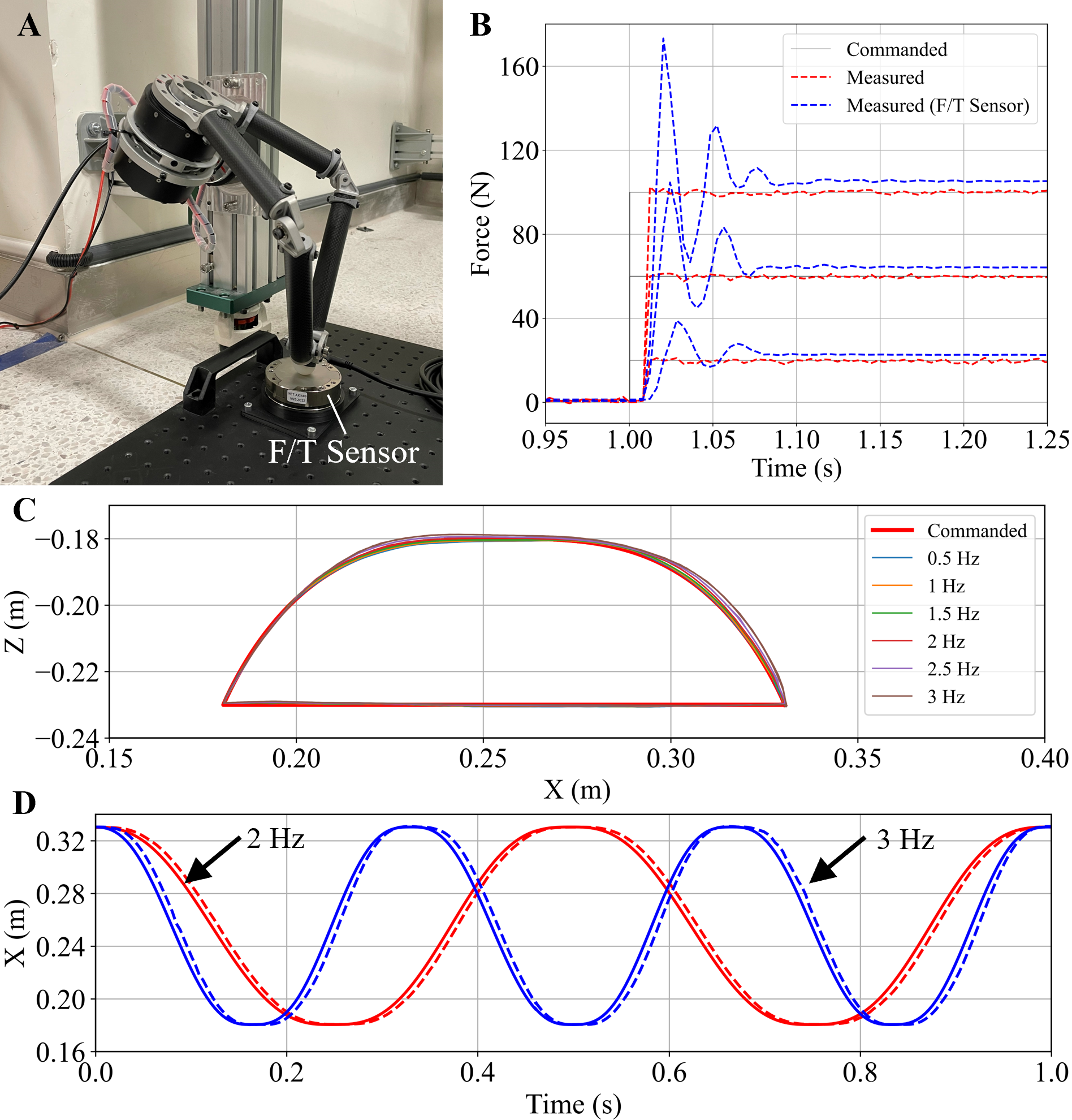}
        \caption{
        \textbf{Experimental results of leg control.}
        (A) Experimental setup to force tracking control. 
        (B) Step response results of 20 N, 60 N, and 100 N input. 
        (C) Position tracking results of the leg for frequencies from 0.5 to 3 Hz. 
        (D) X-direction position tracking results at 2 Hz and 3 Hz trajectories.
        }
        \label{fig:Limb_PosTracking}
    \end{wrapfigure}

\subsubsection{Z-axis Force Tracking}

    As illustrated in Fig.~\ref{fig:Limb_PosTracking}A, an F/T sensor (ATI Axia80-M20, SI-200-8 calibration) is mounted on the ground such that the footsteps on top of the F/T sensor. The six-axis F/T sensor can measure up to 360 N in the $z$-axis with 0.1 N resolution at a sample rate of 1 kHz. Under an assumption of static loading conditions, given the Jacobian, $J \in \mathbb{R}^{3 \times 3}$ and the torque feedback from actuators, $\tau \in \mathbb{R}^{3 \times 1}$, the foot force can also be estimated by 
     \begin{equation}
        f = J^{-T} \tau.
        \label{eq:tau2f}
    \end{equation}
    
    For Cartesian space force control, an open-loop Jacobian transpose control is adopted. The leg configuration is set to the nominal stand pose. To evaluate the force controllability of the leg, the step inputs with 20 N, 60 N, and 100 N are applied, and the results are shown in Fig.~\ref{fig:Limb_PosTracking}B. The solid grey line is the control input. The blue and red dashed lines are the measured force from the F/T sensor and motor current, respectively. The leg system shows reasonable force controllability, and the steady-state errors are about 11.8\%, 6\%, and 5\% of 20 N, 60 N, and 100 N input. The response bandwidth of the leg can be evaluated based on the measured rising time by assuming a second-order system~\cite{Wensing2017Proprioceptive}. In the 100 N step response experiment, the rise time is 8 ms and 9 ms, measured by F/T sensor and motor sensing, resulting in force control bandwidth of 43.75 Hz and 38.8 Hz, respectively.

\subsubsection{Position Tracking}

    The leg is fixed to the aluminum frame without any payload on foot. For position control, a joint PD control with gravity compensation is implemented. A walking trajectory with 0.15 m step length and 0.05 m ground clearance under various frequency ranges (0.5 - 3 Hz) is tested. It is equivalent to the walking speed of 0.15 m/s to 0.9 m/s in trot gait. The foot position is estimated by forward kinematics. Fig.~\ref{fig:Limb_PosTracking}C shows the results of position tracking in Cartesian space. The actual foot position follows accurately along the desired trajectory. The foot's X-axis coordinates tracking at 2 Hz (red) and 3 Hz (blue) are shown in Fig.~\ref{fig:Limb_PosTracking}D. The solid lines represent the desired values, and the dashed lines represent the measured values. The maximum tracking errors at 2 Hz and 3 Hz are 5.6\% and 9\%, respectively.

\subsection{Locomotion Performance Evaluation}

    Theoretical analysis and evaluation of the leg mechanism are essential to give full play to the strengths of the mechanism. For omnidirectional locomotion, the ability to move in all directions, that is, isotropy, should be considered. The feet of the robotic legs need to move fast, whether moving to the foothold in the swing phase or generating reverse speed comparable to body speed in the stance phase. On the other hand, the legs need to produce the vertical force for body weight support and the horizontal force for walking, which puts forward requirements on the payload performance. Thus, we consider the isotropy, the velocity performance, and the payload performance to characterize different leg mechanisms. 

    The isotropy characterizes the similarity of the performance in all directions and how close the mechanism is to the singularity. The inverse condition number based on the 2-norm~\cite{Olds2015Global} is selected to quantify isotropy
    \begin{equation}
        \kappa^{-1} = \frac{1}{\Vert J \Vert_2 \Vert J^{-1} \Vert_2} = \frac{\sigma_{min}}{\sigma_{max}},
    \end{equation}
    where inverse condition number $\kappa^{-1} \in [0,1]$, $\sigma_{min}$, and $\sigma_{max}$ are the minimum and maximum singular values of Jacobian, respectively. 

    \begin{wrapfigure}{r}{0.65\textwidth}
        \centering
        \includegraphics[width=1\linewidth]{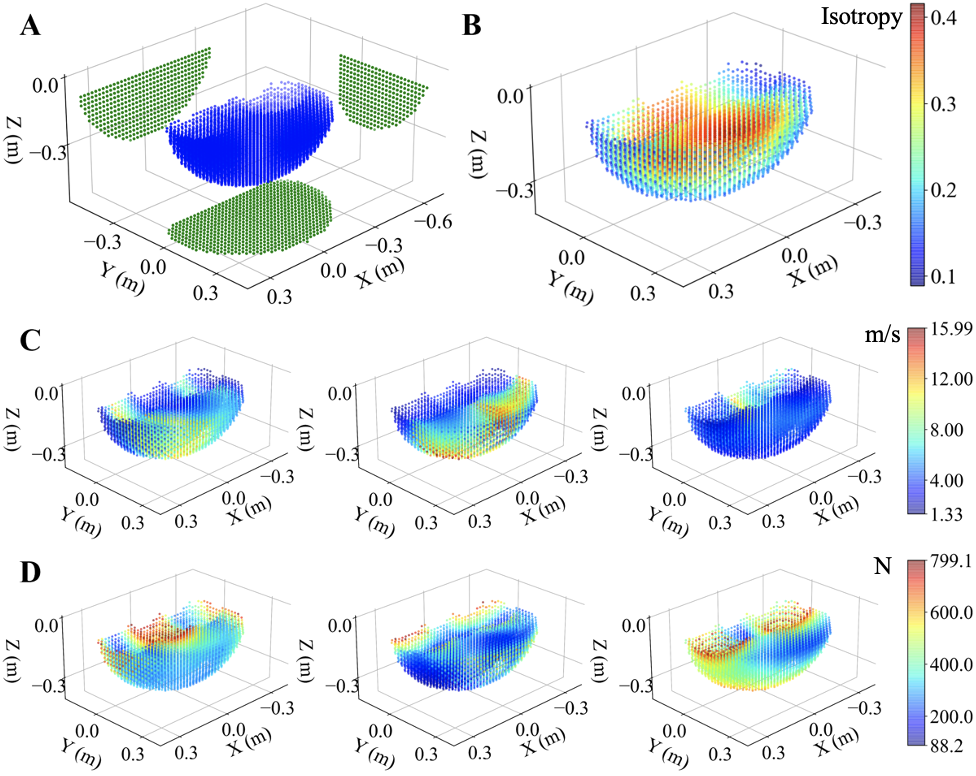}
        \caption{
        \textbf{The locomotion performance evaluation of ORL.}
        (A) The ORL workspace. Blue points represent the position in the workspace, and green points represent the projections in each direction. 
        (B) The point cloud of the isotropic index. 
        (C) The point cloud of the velocity performance index in the X, Y, and Z Directions (from left to right). 
        (D) The point cloud of the payload performance index of X, Y, and Z-direction (from left to right).}
        \label{fig:Limb_Workspace}
        \vspace{-0.5cm}
    \end{wrapfigure}
        
    We choose the maximum foot velocity in a specific direction under the constraint of joint angular velocity to quantify the velocity performance. From differential kinematics, we know $\Dot{q} = J^{-1} v \leq \Dot{q}_{max}$. Considering a unit directional vector $e$, the maximum foot velocity can be calculated by 
    \begin{equation}
        |v_{e, max}| = \min \left( \frac{\Dot{q}_{i,max}}{|J^{-1} e|_i} \right),
    \end{equation}
    where $|v_{e, max}| \in  \mathbb{R}$ is the maximum foot velocity in the direction defined by $e$, and satisfy $v_{e, max} = |v_{e, max}| e$. The $\Dot{q}_{i,max} \in  \mathbb{R}$ is the maximum angular velocity of the $i$th joint, and $|J^{-1} e|_i \in  \mathbb{R}$ is the $i$th element of vector.

    The payload performance index is defined as the maximum foot force. During walking, the quadruped has multiple legs touching the ground, and the horizontal force of each leg can cancel each other out. When computing force in one direction, zero force is not required in other directions. Thus, the payload performance index in each direction can be computed based on Eq. \eqref{eq:tau2f}
    \begin{equation}
        |f_{e, max}| = \sum_{i=1}^3 |e^T J^{-T}|_i \tau_{i,max},
    \end{equation}
    where $|f_{e, max}| \in  \mathbb{R}$ is the maximum foot force in the direction defined by $e$, satisfying $f_{e, max} = |f_{e, max}| e$. The $\Dot{\tau}_{i,max} \in  \mathbb{R}$ is the maximum torque of the $i$th joint, and $|e^T J^{-T}|_i \in  \mathbb{R}$ is the $i$th element of the vector.
    
    The performance indices of ORL are evaluated for the foot positions sampled at 0.02m on the workspace, as shown in Fig.~\ref{fig:Limb_Workspace}. Since these performance indices are configuration-dependent, the Global Performance Index (GPI) defined below is chosen to evaluate leg performance
    \begin{equation}
        GPI = \frac{\int_W c_{PI}\, dW}{\int_W\, dW},
    \end{equation}
    where $c_{PI}$ is the performance index of a specific position, and $W$ is defined as a cubic workspace ($ -0.15 \leq x \leq 0.15, -0.1 \leq y \leq 0.2, -0.3 \leq z \leq -0.15$) that is often utilized and far away from singularities. 

    \begin{wraptable}{r}{0.5\textwidth}
        \vspace{-0.2cm}
        \centering
        \caption{\textbf{Performance comparison of limb mechanisms.}}
        \label{tab:MPC_ComparePerformance}
        \resizebox{1\linewidth}{!}{%
        \begin{tabular}{cccc}
            \hline
            \multicolumn{2}{c}{\textbf{Performance}}                  & \textbf{ORL}   & \textbf{Planar Limb} \\ \hline
            \multicolumn{2}{c}{Isotropy}                     & 0.343 & 0.336            \\ \cline{1-4} 
            \multirow{3}{*}{Velocity performance (m/s)} & $v_x$ & 6.26  & 4.80             \\ \cline{2-4} 
                                                        & $v_y$ & 7.11  & 4.94             \\ \cline{2-4} 
                                                        & $v_z$ & 5.58  & 2.87             \\ \cline{1-4} 
            \multirow{3}{*}{Payload performance (N)}    & $f_x$ & 391   & 237              \\ \cline{2-4} 
                                                        & $f_y$ & 200   & 268              \\ \cline{2-4} 
                                                        & $f_z$ & 377   & 424              \\ \hline
        \end{tabular}%
        }
        \vspace{-0.2cm}
    \end{wraptable}
    As reported in Table~\ref{tab:MPC_ComparePerformance}, the global isotropy performance of ORL is 0.343. The region with high isotropy is in the middle of the workspace, the usual workspace for ORL. The global velocity performance of X, Y, and Z directions are 6.26 m/s, 7.11 m/s, and 5.58 m/s, respectively. It shows that the foot velocity of ORL is faster in the horizontal direction, mainly sideways. Regarding payload performance, the global performance of X, Y, and Z-directions are 391 N, 200 N, and 377 N. The vertical payload capacity of a single leg is 2.7 times the robot's weight.

    Diverse velocities and foothold locations impact leg configuration during locomotion, directly affecting leg velocity and payload performance. Table~\ref{tab:MPC_ComparePerformance} compares the global isotropy, velocity, and payload performance between the ORL and the planar 4-bar leg, with parameters referenced from MIT Cheetah 3~\cite{Bledt2018Cheetah} and TITAN-XIII~\cite{Kitano2016Titan}. Both legs are normalized by total leg length, aligning with the ORL's dimensions. The ORL excels in velocity performance, with a significant 30\%, 44\%, and 94\% enhancement in X, Y, and Z-direction velocities. This discrepancy in leg performance influences quadruped energy efficiency, highlighting the ORL's potential for high-speed locomotion without compromising other key aspects.

\section{Model Predictive Control for Overconstrained Locomotion}
\label{sec:MPC}

\subsection{Model Predictive Control}
    
    To find the optimal control input, we consider a model predictive control problem with $n$ horizon that tracks the CoM reference trajectory, and its standard form can be written as 
    \begin{align}
        \min_{x,u} \quad & \sum_{k=0}^{n-1} ||x_{k+1} - x_{k+1,d}||_{Q_k} + ||u_k||_{R_k}, \\
        s.t. \quad & x_{k+1} = A_k x_k + B_k u_k, \\
                   & \underline{c}_k \leq C_k u_k \leq \overline{c}_k, 
    \end{align}
    where $Q_k$ and $R_k$ are diagonal positive semidefinite matrices that describe the weights of each state and input, $C_k$, $\underline{c}_k$, and $\overline{c}_k$ is inequality constraints of the control input which is usually the friction cone constraints.
    
    The future state can be computed ahead of time by the following equation
    \begin{equation}
        x_n = \left( \prod_{k=0}^{n-1}A_k \right) x_0 + 
        \sum_{k=0}^{n-2} \left[ \left( \prod_{i=k+1}^{n-1} A_i \right) B_k u_k \right]
        + B_{n-1} u_{n-1}.
    \end{equation}
    Then we can define the state sequence as $X = [x_1^T, x_2^T, \ldots , x_n^T]^T$ and the control input sequence as $U = [u_0^T, u_1^T, \ldots , u_{n-1}^T]^T$. The condensed formulation of the dynamics during the prediction horizon is
    \begin{equation}
        X = A_{qp} x_0 + B_{qp} U.
    \end{equation}
    The objective function of the model predictive control problem in condensed form is 
    \begin{equation}
        J(U) = (A_{qp} x_0 + B_{qp} U - X_d)^T Q (A_{qp} x_0 + B_{qp} U - X_d) + U^T R U,
    \end{equation}
    where $X \in \mathbb{R}^{13h \times 1}$ is the vector of all states, $A_{qp} \in \mathbb{R}^{13h \times 13}$ is the state matrix, $x_0 \in \mathbb{R}^{13 \times 1}$ is the current state, $B_{qp} \in \mathbb{R}^{13h \times 12h}$ is the input matrix, $U \in \mathbb{R}^{12h \times 1}$ is the vector of all inputs, $Q \in \mathbb{R}^{13h \times 13h}$ is the diagonal matrix of weights for states, and $R \in \mathbb{R}^{12h \times 12h}$ is the diagonal matrix of weights for control inputs. Finally, the model predictive control problem becomes a quadratic programming problem with linear constraints that can be solved in real-time. In this paper, a model predictive problem with ten horizons can be solved at 250 Hz using the C++ and OSQP~\cite{Stellato2020Osqp} quadratic programming library.

\subsection{Simulating Overconstrained Locomotion using MPC}

\begin{figure}[htbp]
        \centering
        \includegraphics[width=0.9\linewidth]{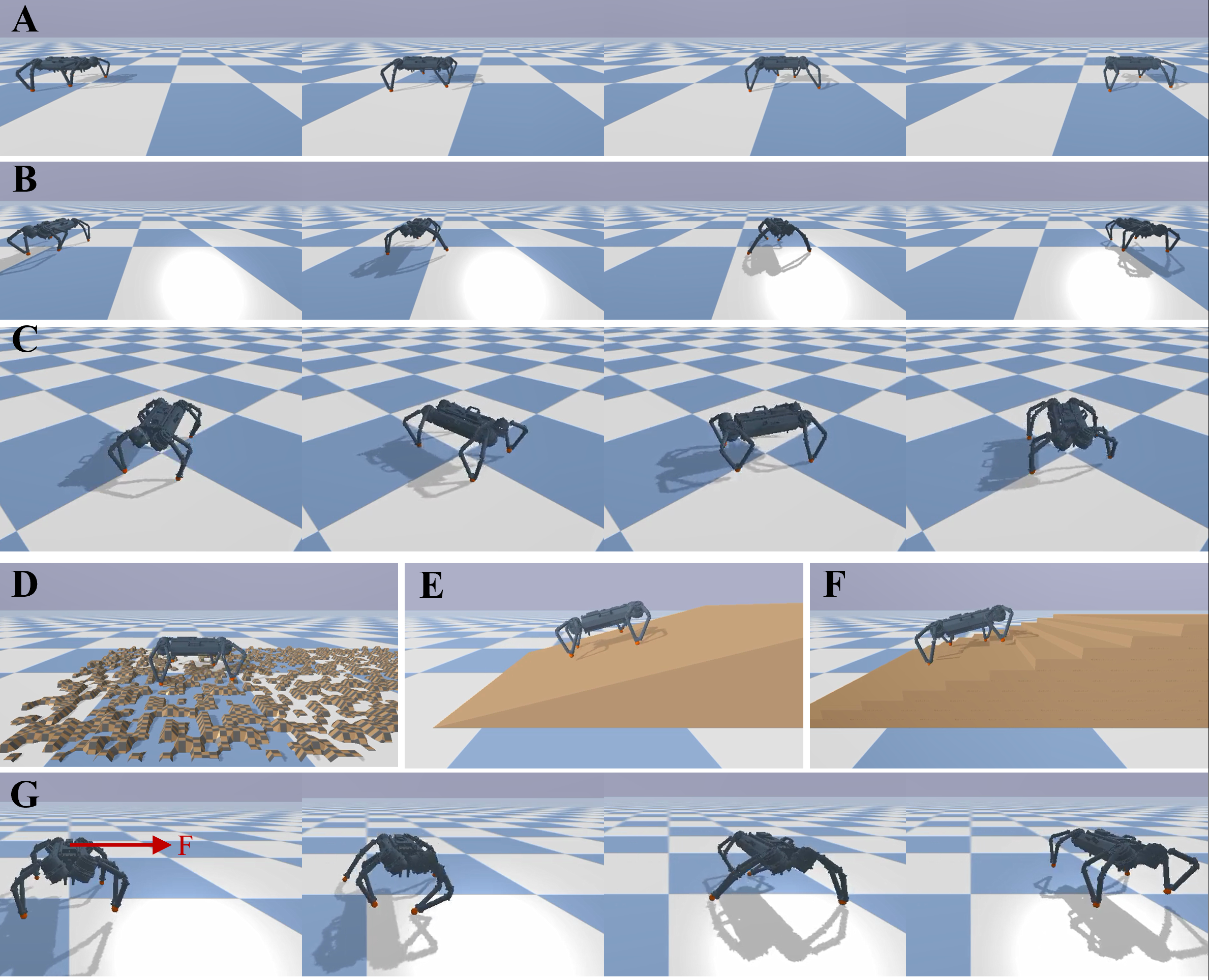}
        \caption{
        \textbf{Snapshots of simulated overconstrained locomotion using MPC.}
        (A) Forward trotting. 
        (B) lateral trotting. 
        (C) Turning on the spot. 
        (D) Trotting on gravel terrain. 
        (E) Trotting on the slope. 
        (F) Trotting on the stairs. 
        (G) Recovery from a lateral push.
        }
        \label{fig:MPC_PybulletSims}
    \end{figure}

    To validate the effectiveness of the proposed methods, we conducted various locomotion simulations, including omnidirectional locomotion, uneven terrain locomotion, and push recovery, as shown in Fig.~\ref{fig:MPC_PybulletSims}. 
    The simulations were performed in Pybullet with a control bandwidth of 250 Hz, the same as the physical prototype. The URDF model is generated from the CAD model that is as close as possible to the physical prototype. The trotting gait with a duty factor of 0.5 and a gait period of 0.5 s is adopted in all simulations. The model predictive controller is implemented with ten horizon steps and can run at 250 Hz for 0.2 s prediction. The controller gain of each step are set $Q_k = diag(10,10,0.01, 2,5,50, 0.01,0.01,2, 0.1,0.4,0.4, 0)$ and $R_k = 0.0001 \mathbb{I}_{12}$.

\subsubsection{Omnidirectional Locomotion}
    
    In the omnidirectional trotting simulation, the overconstrained quadruped tracks the velocity commands for forward, lateral, and turn on the spot separately, as shown in Figs.~\ref{fig:MPC_PybulletSims}A-C. 

\subsubsection{Gravel Terrain}

    A simulation of gravel terrain traveling was carried out to verify the controller for unknown terrain adaptability. The terrain elevation is randomly distributed within a maximum height of 0.1 m (25\% of the total leg length). As shown in Fig.~\ref{fig:MPC_PybulletSims}D, the overconstrained quadruped can smoothly traverse the gravel terrain at a velocity of 0.5 m/s. Although there would be a slight slip when touching the gravel, the controller can stabilize the posture soon.

\subsubsection{Sloped/Stairs Terrain}

    By estimating the slope, the overconstrained quadruped can be commanded to be parallel to the slope and trotting on the slope. In simulations, the robot trots at 0.5 m/s on a 15${}^{\circ}$ slope or the stairs with a height of 0.05 m and a length of 0.2 m, as shown in Figs.~\ref{fig:MPC_PybulletSims}E and F. The robot can maintain a stable posture and hardly slips. During trotting on sloped terrain and stairs, the friction cone constraints of the model predictive controller need to be modified accordingly to accurately reflect the position of the contact plane relative to the world coordinate system.
    
    \begin{figure*}[htbp]
        \centering
        \includegraphics[width=1.0\linewidth]{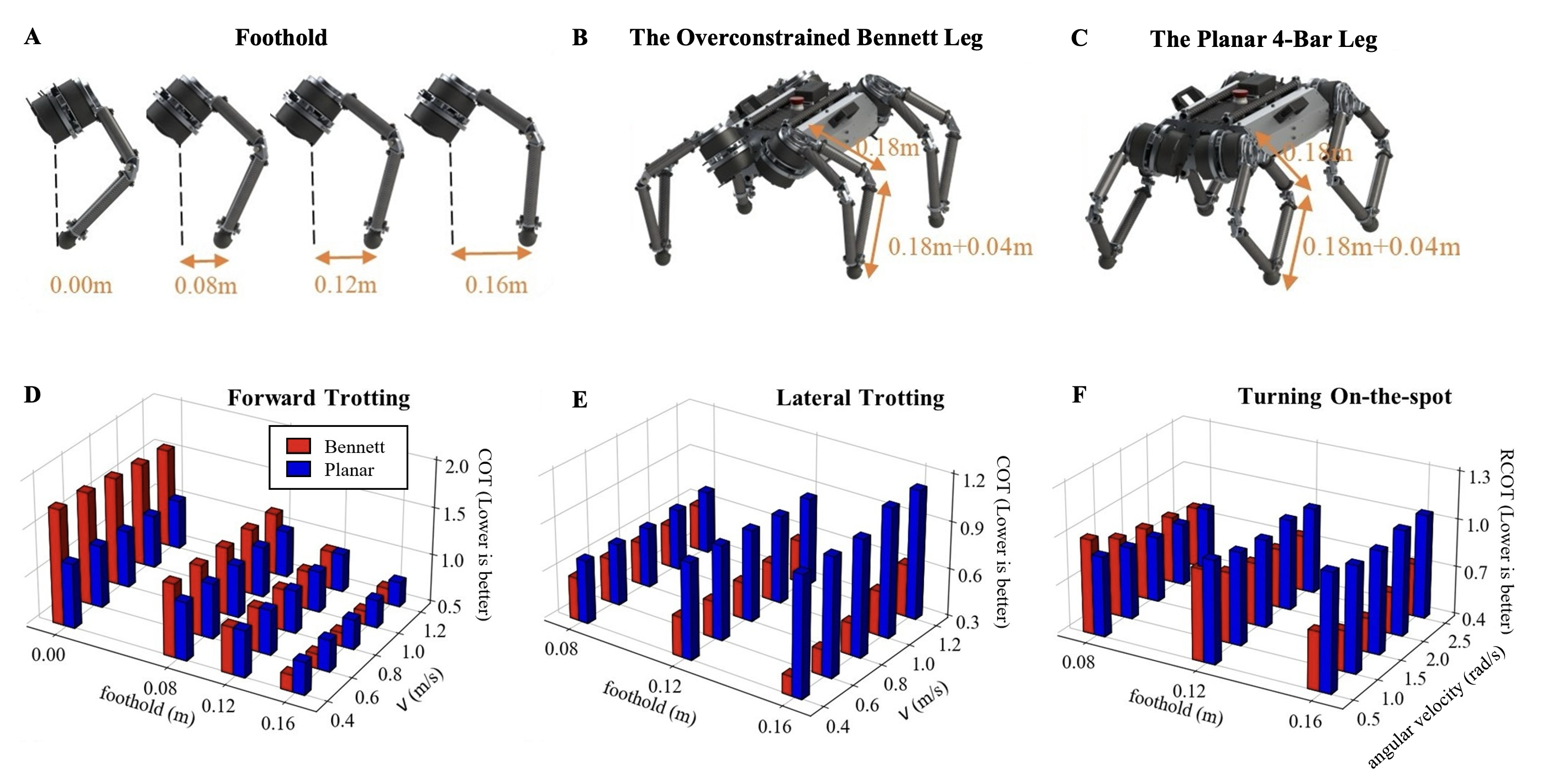}
        \caption{
        \textbf{The COT results of the ORL quadruped (red) and planar 4\textit{R} quadruped (blue).} 
        (A) The schematic diagram of different footholds. 
        (B-C) The leg parameters of the overconstrained quadruped and the quadruped with planar 4-bar linkage.
        (D) The COT of forward trotting with different velocities and footholds. 
        (E) The COT of lateral trotting with different velocities and footholds. 
        (F) The RCOT of turning with different velocities and footholds. 
        COTs of 0.0m footholds for lateral trotting and turning-on-the-spot are not presented as they lead to unstable gaits.
        }
        \label{fig:MPC_CompareCoT}
    \end{figure*}
        
\subsubsection{Push Recovery}

    The lateral stability of the quadruped robot is generally weaker than the forward stability, and a simulation of lateral push recovery is conducted to verify the effectiveness of the controller. A lateral force of 140 N (equivalent to the robot's weight) is applied at the CoM and lasts for 0.2 s. The change of lateral velocity is up to 1.8 m/s, and the pitch change is about 0.1 rad, but it can still stabilize within 2 s.

\subsection{Energy Efficiency Analysis}

    The locomotion efficiency of the quadruped is affected by factors such as gait, leg mechanism, velocity, foothold location, and so on. We carried out a series of simulations under trot gaits in different footholds shown in Fig.~\ref{fig:MPC_CompareCoT}A. 
    
    The proposed ORL in Fig.~\ref{fig:MPC_CompareCoT}B is compared with the planar 4-bar leg in Fig.~\ref{fig:MPC_CompareCoT}C. The link lengths of these two legs remain the same. Only the twist angles are different. The twist angle of ORL and that of the planar 4-bar leg are 120${}^{\circ}$ and 180${}^{\circ}$, respectively. Secondly, three basic motions are evaluated, including forward trotting, lateral trotting, and turning on the spot. In addition, the neutral point, which affects foothold location, is also considered. Finally, different trotting velocities ranging from 0.4 to 1.2 m/s are tested.

    The cost of transport (COT) is chosen to characterize the efficiency of quadrupedal locomotion. The smaller the COT, the higher the efficiency. The COT is calculated through the mechanical energy of all motors 
    \begin{equation}
        COT = \frac{P}{mgv} = \frac{\sum_{j=1}^{12} \int |\tau_{j} \omega_{j}| dt}{mgd},
    \end{equation}    
    where $P$ is the power, $d$ is the locomotion distance, $\tau_j$ and $\omega_j$ is the torque and velocity of $j{th}$ motor. The rotational cost of transport (RCOT) is defined as the COT corresponding to the velocity of the far-end point of the body to represent the energy consumption of turning on the spot, and the formula is expressed as
    \begin{equation}
        RCOT = \frac{P}{mg \omega_z \frac{\sqrt{L^2 + W^2}}{2}},
    \end{equation}
    where $\omega_z$ is the angular velocity of quadruped, $L$ and $W$ is the body length and body width.

    The simulation results of the forward trotting, lateral trotting, and turning of the quadrupeds with different limb designs are shown in Figs.~\ref{fig:MPC_CompareCoT}D-F, respectively. The results show that:
    \begin{enumerate}
        \item With the increase of foothold, the COT of forward trotting gradually decreases. Generally, the selection of foothold location is related to the hip position. When the foothold is 0, the torque and velocity of the hip joint are almost zero, but the COT does not decrease and is even higher.
        \item When the foothold is 0.16 in forward trotting and turning, the overconstrained quadruped's COT is lower than the quadruped with the planar 4-bar leg.
        \item For lateral trotting, the overconstrained quadruped's COT is always much lower than the quadruped with the planar 4-bar leg, saving up to 60\% of the energy cost (velocity is 0.4 m/s and foothold is 0.16 m).
        \item For the overconstrained quadruped, the energy efficiency in lateral trotting is better than forward trotting, and the best case can reach 55\% improvement (velocity is 0.4 m/s and foothold is 0.08 m).
        \item Compared with quadruped with planar 4-bar linkage, the COT of the overconstrained quadruped is less sensitive to the velocity of forward trotting but more sensitive to the velocity of lateral trotting.
    \end{enumerate}

\section{Reinforcement Learning for Overconstrained Locomotion}
\label{sec:RL}

\subsection{Formulation of Reinforcement Learning}
    
    Reinforcement learning has been a practical framework for training quadruped robots to navigate diverse terrains with agility and efficiency~\cite{Tan2018Sim, Lee2020Learning}. We also trained the overconstrained locomotion using the latest robot learning framework Orbit~\cite{mittal2023orbit} built upon NVIDIA Isaac Sim. The locomotion task is to follow a specified velocity and heading direction.

    The observation space of flat-terrain trotting consists of base linear and angular velocities, commanded base linear and angular velocities, joint positions and velocities, projected gravity, and the last applied actions. For multi-terrain RL, the observation space consists of all the observations in the flat-terrain task and measured terrain heights sampled around the robot's base. The action space is the joint positions of the 12 actuators, and each action of desired joint positions is converted to motor torques using a PD controller.

    We keep most reward functions the same as Orbit's Unitree A1 robot training environment, such as linear and angular velocity tracking rewards, action rate penalties, and penalties on velocities in undesired directions. To optimize the energy efficiency of overconstrained locomotion, we include an extra actuator energy penalty term calculated from the following formula at each timestep:
    \begin{equation}
        R_{energy} = \sum_{j=1}^{12} \Delta\theta_{j}\tau_{j},
    \end{equation}
    where $\Delta\theta_{j}$ is the change of joint position, $\tau_{j}$ is the applied torque of $j_{th}$ motor. We use the PPO algorithm~\cite{Schulman2017PPO} to optimize the locomotion policy. Please refer to the supplementary material for detailed reward functions and PPO hyperparameters.

\subsection{Overconstrained Locomotion learned from RL}

\subsubsection{Footholds on Flat Terrain}

\begin{figure}[htbp]
        \centering
        \includegraphics[width=0.75\linewidth]{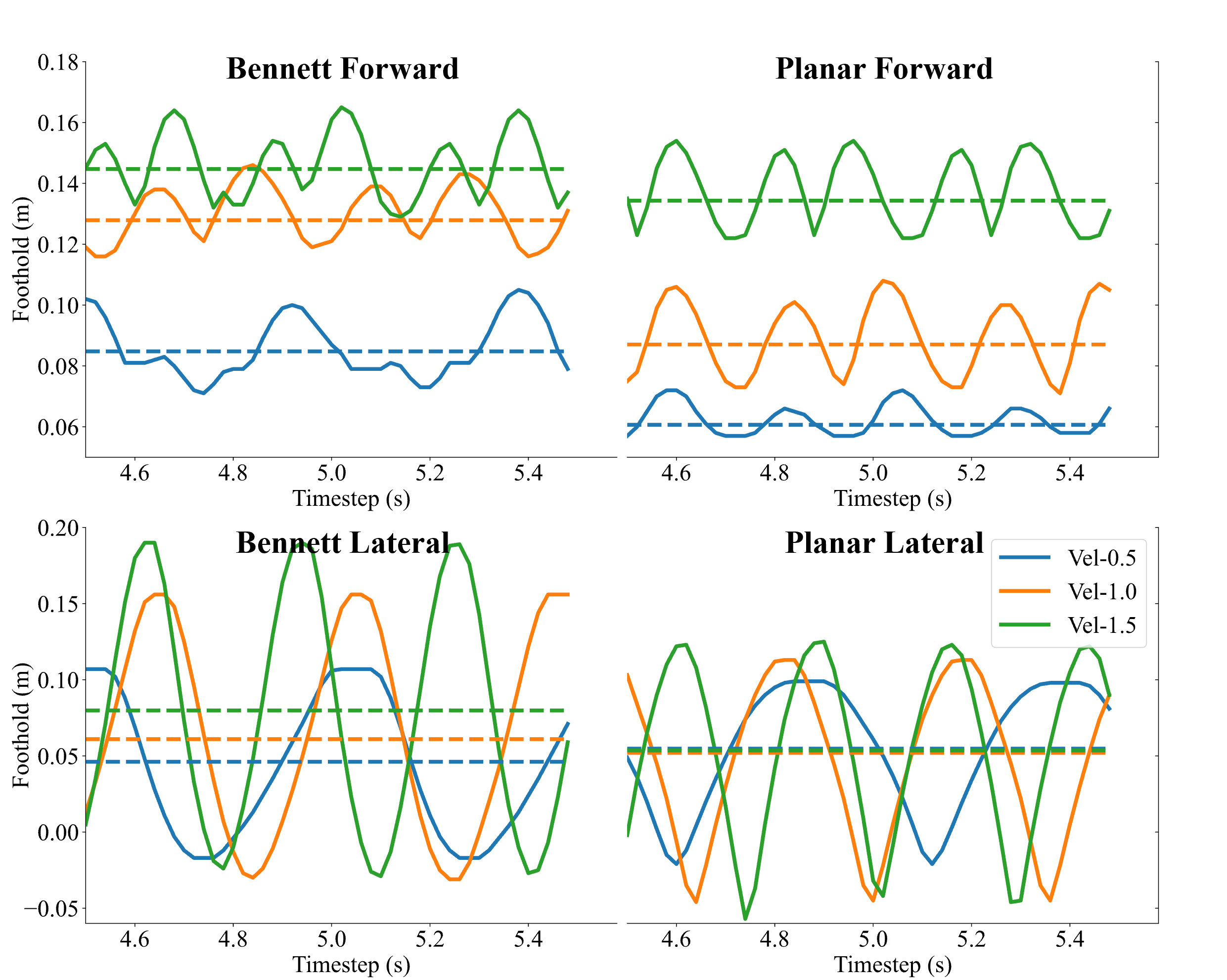}
        \caption{
        \textbf{The real-time foothold of locomotion policies learned from reinforcement learning.} Time series of footholds in forward (top) and lateral (bottom) trotting for robots with Bennett and planar linkages, respectively. The dashed lines are the average footholds.
        }
        \label{fig:RL_Foothold}
    \end{figure}

    To investigate the gaits of overconstrained locomotion in flat terrain and facilitate the efficiency analysis, we trained 4 locomotion policies optimized for Bennett and planar leg designs in forward and lateral trotting tasks. To make the comparison fair, we performed a grid search of various PPO hyperparameter combinations for each task and selected the one with the lowest energy penalty. For each of the four policies, We let the robot trot at the speed of 0.5 m/s, 1.0 m/s, and 1.5 m/s for 10 seconds and recorded the real-time foothold position. The foothold position becomes periodical after the robot reaches the commanded velocity, and a short period of the foothold position for each policy is shown in Fig.~\ref{fig:RL_Foothold}.

    We found that in forward trotting, the foothold increases as the commanded velocity increases for both robots. With higher velocity, the energy cost of the joints also increases, and the policy has to optimize the energy penalty term more to gain more rewards. The policy naturally learns how to enhance energy efficiency by increasing the foothold, consistent with our findings in Fig.~\ref{fig:MPC_CompareCoT}D. In lateral trotting, the Bennett quadruped is also found to increase its foothold as the commanded velocity increases, which agrees well with the COT result from the MPC controller shown in Fig.~\ref{fig:MPC_CompareCoT}E that a larger foothold leads to lower COT. The planar quadruped's foothold in lateral trotting remains unchanged instead of decreasing, as indicated by the MPC results that a smaller foothold leads to lower COT for a planar quadruped. The reason is that the smaller foothold is unstable and prone to falling, and the unchanged foothold results from balancing the two rewards competing against each other. These results align with literature published recently~\cite{He2024AgileSafe}, where a wider foothold supports agile and safe locomotion at high speed.

\subsubsection{Multi-terrain}

\begin{figure}[htbp]
        \centering
        \includegraphics[width=0.9\linewidth]{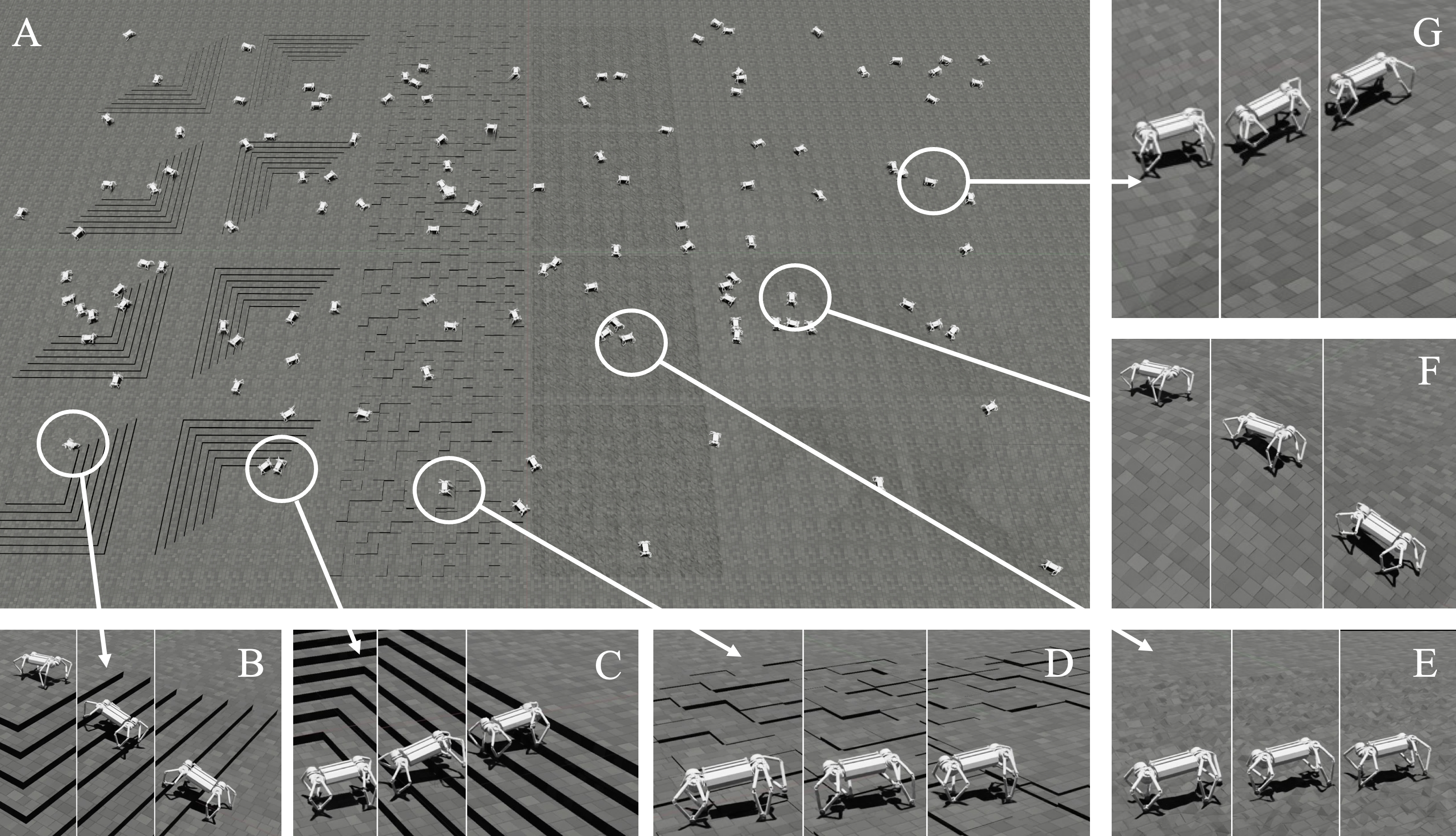}
        \caption{
        \textbf{Snapshots of simulated overconstrained locomotion using reinforcement learning trained on multi-terrain.}
        (A) Training scene on multi-terrain. 
        (B) Pyramid terrain.
        (C) Inverted pyramid terrain.
        (D) Discrete random height fields.
        (E) Uniform random height fields.
        (F) Pyramid slope.
        (G) Inverted pyramid slope.
        }
        \label{fig:RL_IssacSims}
    \end{figure}

    Our multi-terrain training environment was adapted from the framework described by Rudin et al.~\cite{rudin2022learning}, with specific modifications to accommodate our proposed overconstrained quadruped design. We developed this learning environment by adjusting the locomotion environment available in Orbit, as depicted in Fig.~\ref{fig:RL_IssacSims}A. 
    We used the Unified Robot Description Format (URDF) to construct our robot's simulation model, seamlessly exported via ACDC4Robot~\cite{Qiu2023ACDC4Robot}. ACDC4Robot is a tool specifically engineered to streamline the conversion of design models into simulation-ready formats well-suited for learning environments. The training reward converges to a high value, and the overconstrained robot successfully transverses the designed terrains after training, as shown in Fig.~\ref{fig:RL_IssacSims}B-G.

    \begin{figure}[htbp]
        \centering
        \includegraphics[width=0.9\linewidth]{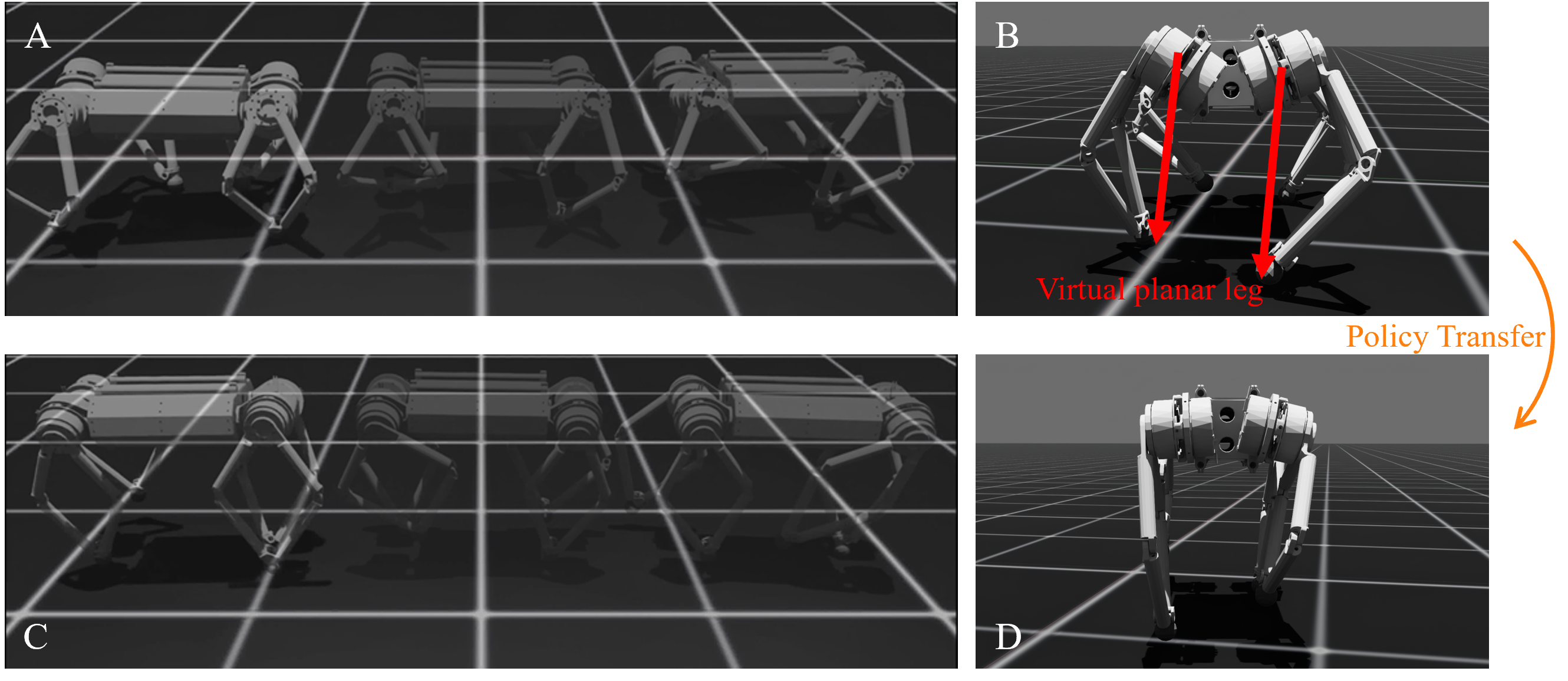}
        \caption{
        \textbf{Transfer of overconstrained locomotion policy to the quadruped robot with planar linkages.}
        (A) Snapshots of forward trotting of learned overconstrained policy.
        (B) Front view of the overconstrained quadruped during trotting.
        (C) Snapshots of planar quadruped forward trotting using overconstrained policy.
        (D) Front view of the planar quadruped during trotting.
        }
        \label{fig:RL_IssacSims_Transfer}
    \end{figure}  

    In our current settings, the overconstrained quadruped robot tends to adopt a gait that positions the foothold near its centroid. Although this gait might not be the most energy-efficient, it is easier for the robot to overcome challenging terrains as a more significant foothold might lower the height of the robot, which will make the robot base frequently collide with the terrain. This learned policy demonstrates remarkable transferability to robots with planar linkages, mainly on flat surfaces, as illustrated in Figs.~\ref{fig:RL_IssacSims_Transfer}A\&C. The possible reason is that the overconstrained policy's effective virtual planar legs (Fig.~\ref{fig:RL_IssacSims_Transfer}B) share similarities to the legs of the planar robot despite the distinct differences in their linkage mechanisms. This observation suggests the potential for developing a universal policy applicable across different robotic leg designs. Moving forward, we aim to delve into the intricate relationships between various mechanism designs and limb skills, which could unveil new avenues for optimization techniques in robotic locomotion.  

\subsection{Energy Efficiency Analysis}
    
\begin{figure}[htbp]
        \centering
        \includegraphics[width=0.6\linewidth]{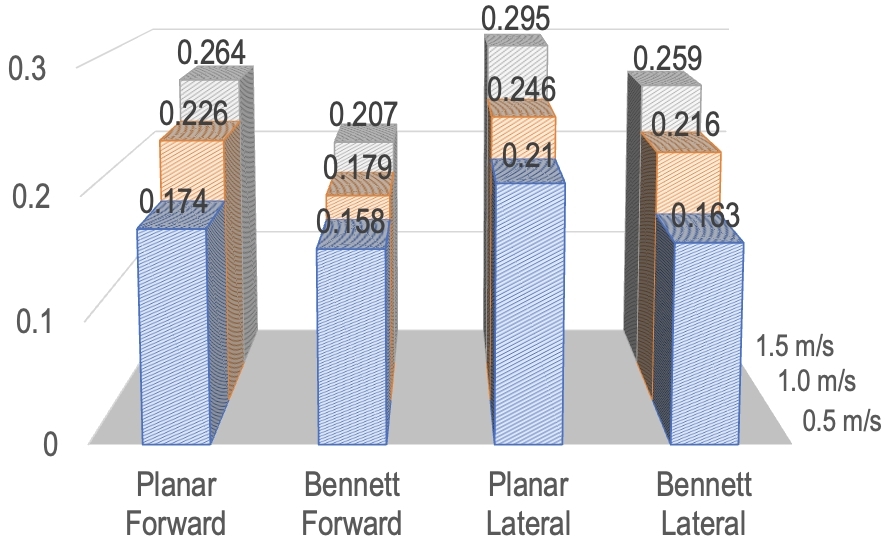}
        \caption{
        \textbf{The COT results of the ORL quadruped and planar 4\textit{R} quadruped.}
        Four locomotion policies are trained for the two quadrupeds in forward and lateral trotting. Each column shows the COT of one policy under different speeds.
        }
        \label{fig:RL_CompareCoT}
    \end{figure}

    We further benchmark the COTs of forward and lateral trotting of the quadrupeds with different limb designs under their respective optimal locomotion policies obtained from full-model optimization in RL. Note these results may not directly correspond to the results in MPC simulation as the trotting speed determines the foothold in an RL policy. Hence, the COT results here do not have an independent dimension of foothold. As shown in Fig.~\ref{fig:RL_CompareCoT}, we have the following observations.
    \begin{enumerate}
        \item The COT of the overconstrained quadruped is always lower than the planar quadruped when trotting at the same speed on flat terrain. 
        \item In forward trotting, energy cost reduction brought by Bennett limbs is more prominent for faster walking speed, saving 9\% at 0.5 m/s and 22\% at 1.5 m/s.
        \item For both overconstrained and planar quadrupeds, forward trotting could achieve higher energy efficiency than lateral trotting at different speeds with their optimal locomotion policy, respectively. 
        \item In forward trotting, the overconstrained quadruped's COT is more unanimous under different velocities (standard deviation of 0.02) compared to the planar quadruped (SD of 0.05). On the contrary, the COT of the overconstrained quadruped is more sensitive to velocity in lateral trotting. This agrees well with the results using MPC.
    \end{enumerate}
    
\section{Conclusion, Limitations, and Future Work}
\label{sec:Conclude}

    The overconstrained linkages have contributed significantly to the literature on robotics due to their kinematic behaviors~\cite{Mavroidis1995Analysis, Waldron1973StudyI} and singularity reconfiguration~\cite{Wang1987AStudy}, providing a rich pool of advanced mechanisms for developing various mathematical tools used widely in applied robotics~\cite{Pennock2018Professor, Raghavan1990Kinematic, Waldron1973StudyII}. However, their applications in modern robotics are still limited due to the complex spatial motions and restrictive geometric constraints~\cite{Dai2012Advances}. In this work, we proposed a modular limb, driven by quasi-direct drives, parametrically reconfigurable between a planar and a Bennett linkage, closely resembling their biologically inspired roots for legged motion. We implemented model predictive control with simulated locomotion in various tasks. Inspired by ETH's research on training ANYmal in a multi-terrain scenario, we also successfully implemented reinforcement learning of the overconstrained locomotion in simulation. Simulated MPC results show that the overconstrained robotic limbs exhibit a superior Cost-of-Transport in omnidirectional tasks, including forward trotting, lateral trotting, and turning on the spot, compared to the planar limb design when the footholds distance is taken into consideration, which characterizes the biological difference between the reptile and mammal animal limbs. This superior energy efficiency of overconstrained robotic limbs is further verified in locomotion policies optimized in reinforcement learning. This is the first work in literature to report the superior performance of overconstrained robotic limbs in dynamic locomotions compared to planar ones, which currently dominate the limb design of legged robots.

    We also acknowledge several limitations of this study. While the superior energy efficiency was previously verified in an overconstrained quadruped actuated by servo motors, further verification is needed to fully validate the results reported in this study, where QDDs are used on a full-sized legged robot. For example, theoretically speaking, when executing a reptile-like gait, the overconstrained limbs exhibit a wider foothold that will inevitably induce an extra torque to the hip motor but require less output from the two co-axially arranged motors. This seems to be a more balanced requirement for the motors when they are usually identical in most modern quadruped designs, which needs further verification through hardware implementation. We found that the MPC controller is more stable in overconstrained locomotion, while parameter tuning for reinforcement learning of overconstrained locomotion is more sensitive to changes but more adaptive to different terrain conditions. The quadrupedal prototype used in this study has limited sensory feedback due to the lack of environmental perception devices, and only a single inertial measurement module is used. Further integration is required to improve the robot's control and interactions. Additionally, a dedicated filtering algorithm is needed to process sensor data accurately for closing the Sim2Real gap. 
    
    Nevertheless, the results shown in this paper potentially pave the path to a new field of overconstrained robotics, where the overconstrained geometry of spatial linkages could be applied to designing novel robots to achieve complex movement within a simple design. Engineering optimization on the limb design is also necessary to build the overconstrained quadruped, specifically with the Bennett linkage, without considering its parametric reconfiguration, further simplifying the limb design with fewer components. 

\section*{Acknowledgments}

    This work was partly supported by the National Natural Science Foundation of China [62206119] and the Science, Technology, and Innovation Commission of Shenzhen Municipality [JCYJ20220818100417038], Shenzhen Long-Term Support for Higher Education at SUSTech [20231115141649002], SUSTech Virtual Teaching Lab for Machine Intelligence Design and Learning [Y01331838]. 

\bibliographystyle{unsrt}
\bibliography{References}
\newpage
\section*{Appendix A on Engineering Specifications}
\label{App-A:Specs}

    While recent research established the theoretical advantage of overconstrained robotic limbs for omnidirectional locomotion~\cite{Gu2022Overconstrained}, only servo motors were used with validations through a light-weight robot that is functionally limited in dynamic gait control~\cite{Gu2023Computational}. In this work, we present a generalized robot design actuated by three QDDs per limb, with each limb involving a hip actuation attached horizontally to the body and a co-axial dual-output actuation for the limb mechanism. Note that there is also a possible design with a vertical configuration of the hip actuation using the same limb module~\cite{Sun2023Bridging}, where the overconstrained robotic limb (ORL) is used for object manipulation besides ground locomotion.

    Quasi-direct Drives are generally considered an engineering-balanced choice~\cite{Bledt2018Cheetah, Arm2019Spacebok} for actuating robotic limbs in small and medium-sized quadruped with dynamic locomotion, considering various factors such as back-drivability, torque density, energy efficiency, hardware cost, and proprioceptive sensing when compared to other choices such as high-gear-reduction motors~\cite{Raibert2008Bigdog, Bellicoso2018Advances}, direct drives~\cite{Asada1987Direct, Kenneally2016Design}, and series elastic actuator~\cite{Pratt1995Series}. In our design, for the sake of comparison, we chose an off-the-shelf QDD actuator (Unitree A1 motor) for the overconstrained quadruped, which has a gear reduction ratio of 9.1, a peak output torque of 33.5 Nm, and a peak output speed of 21 rad/s, weighing 605 g.

    \begin{table}[htbp]
        \centering
        \caption{\textbf{Overconstrained quadruped specifications.}}
        \label{tab:EngSpecs}
        \resizebox{0.6\columnwidth}{!}{%
        \begin{tabular}{cc}
        \hline
        \textbf{Property}                 & \textbf{Parameters}                              \\ \hline
        Length $\times$ Width $\times$ Height (m)  & 0.51 $\times$ 0.45 $\times$ 0.32                          \\
        Active DoF number        & 12                                      \\
        Total weight (kg)        & 14                                    \\
        Inertial measurement unit  & \multicolumn{1}{l}{MicroStrain 3DM-GX3} \\
        Computing board          & NUC i7-1260P                            \\
        Joint peak torque (Nm)   & 33.5                                    \\
        Joint peak speed (rad/s) & 21                                      \\ \hline
        \end{tabular}%
        }
    \end{table}

    All three actuators are mounted close to the body, significantly decreasing leg inertia. Each link of the Bennett linkage is constructed by two aluminum alloy hinges and a carbon fiber tube, which are used to fix the twist angles and reduce weight, respectively. An alternative form design (the theoretical axis of links does not coincide with the axis of the carbon fiber tube) of linkage is adopted to enlarge the range of motion. The Bennett linkage has two singular configurations: the angle between two active links $\eta = 0^{\circ}$ and $\eta = 180^{\circ}$. And these two configurations are avoided by mechanical limits. The foot is designed with a hemispherical point foot made of 60A rubber. The links have a total mass of 1.6 kg for all four legs and account for 11\% of the mass of the robot.

    All electrical components, including communication converters, power distributed board (PDB), inertial measurement unit (IMU), battery, and embedded PC, are integrated into the trunk, and those are arranged evenly to keep the center of mass (CoM) close to the geometric center of the trunk. We chose a 6s 5700 mAh LiPo battery with a nominal 24 V voltage and a 670 g mass for the power input. The PDB provides four 24 V lines for the actuators, a 20 V line for the embedded PC, and a 5 V line for the relay. An Intel NUC equipped with an Intel Core i7-1260P @ 4.8 GHz is the embedded PC with RT-Linux. This primary controller solves the control task and communicates in real-time with 12 actuators via high-speed RS485 protocol. The data measured by the IMU (MicroStrain 3DM-GX3-25) are transferred into the embedded PC through USB protocol at 500 Hz.

\vspace{0.5cm}
\section*{Appendix B on Overconstrained Quadruped Modeling}
\label{App-B:Modeling}

\subsection*{Forward Kinematics}

    The forward kinematics of the overconstrained quadruped's ORL consists of two steps:
    \begin{enumerate}
        \item Find the passive joint angles using the geometric constraints of the Bennett linkage.
        \item Solve the forward kinematics of the 3-DoF spatial serial RRR manipulator as an open chain.
    \end{enumerate}
    Based on the explicit enclosure equations of the ORL, the mapping from active joints $q_a = [q_1, q_2, q_4]^T$ to the joints of spanning tree $q = [q_1, q_2, q_3, q_4, q_5]^T$ is obtained
    \begin{equation}
        q = \gamma(q_a) =
        \begin{bmatrix}
         q_1 \\ 
         q_2 \\
         2 \atan{(k t_{\eta})} \\
         q_4 \\ 
         2 \atan{(k t_{\eta})}
        \end{bmatrix},
        \label{eq:qa2q}
    \end{equation}
    where $t_{\eta}$ means the $\tan{\eta}$, \(k = \csc{(\frac{\beta - \alpha}{2})}\), and \(\eta = \frac{q_4 - q_2}{2}\).

    Only one serial chain is considered when calculating the foot's position. Taking the front left leg as an example, we choose the serial chain on the right side ($1 \rightarrow 4 \rightarrow 5 \rightarrow 6$). The screws and home matrix are as below
    \begin{equation}
        \begin{cases}
            & \mathcal{S}_1 = \begin{bmatrix}
                1 & 0 & 0 & 0 & 0 & 0
            \end{bmatrix}^T, \\
            & \mathcal{S}_2 = \begin{bmatrix}
                0 & 1 & 0 & 0 & 0 & 0
            \end{bmatrix}^T, \\
            & \mathcal{S}_3 = \begin{bmatrix}
                s_{\beta} & c_{\beta} & 0 & l_3 c_{\beta} & -l_3 s_{\beta} & -l_1 s_{\beta}
            \end{bmatrix}^T, \\
            & M = \begin{bmatrix}
                1 & 0 & 0 & 0 \\
                0 & 1 & 0 & l_1 \\
                0 & 0 & 1 & -(l_3 + l_2 + l_4) \\
                0 & 0 & 0 & 1
            \end{bmatrix},
        \end{cases}
    \end{equation}
    where $s_{\beta}$ and $c_{\beta}$ stand for $\sin{\beta}$ and $\cos{\beta}$, respectively. $\mathcal{S}_1, \mathcal{S}_2$, and $\mathcal{S}_3$ are the screws of each joint. $M$ is the position and orientation of the foot respective to hip frame $\{H\}$ when all active joint angles are zero. 

    As all the joint angles of the spanning tree model are obtained from Eq. \eqref{eq:qa2q}, the position and orientation of the foot with respective to hip frame $\{H\}$ can be computed by the product of exponential formula
    \begin{equation}
        ^{H}T = e^{[\mathcal{S}_1]q_1} e^{[\mathcal{S}_2]q_{4}} e^{[\mathcal{S}_3]q_{5}} M,
        \label{eq:PoE}
    \end{equation}
    where $^{H}T$ is the homogeneous matrix from hip to foot and $e^{[\mathcal{S}_i]q_j}$ is the homogeneous matrix of each screw in exponential coordinates.

\subsection*{Inverse Kinematics}

   The corresponding inverse kinematics problem also includes two steps:
    \begin{enumerate}
        \item Solve the inverse kinematics of the 3-DoF spatial serial RRR manipulator.
        \item Find the active joint angles by the closure equations of Bennett linkage.
    \end{enumerate}

    Multiplying the inverse of $e^{[\mathcal{S}_1]q_1}$ to the homogeneous matrix of the foot, the active joint variables can be separated. From the $(1,4)$, $(2,4)$, and $(3,4)$ terms, we obtain
    \begin{equation}
        \begin{cases}
            & p_x = -(l_2+l_4) (s_4 c_5 + c_{\beta} c_4 s_5) - l_3 s_4,  \\
            & p_y c_1 + p_z s_1 = l_1 + (l_2 + l_4) s_{\beta} s_5,  \\ 
            & -p_y s_1 + p_z c_1 = -(l_2 + l_4) (c_4 c_5 - c_{\beta} s_4 s_5) - l_3 c_4, 
        \end{cases}
        \label{eq:PoE2_14_24_34}
    \end{equation}
    where $(p_x, p_y, p_z)$ is the position of the foot. Solving the Eq. \eqref{eq:PoE2_14_24_34}, we get
    \begin{equation}
        \begin{cases}
            & A_1 s_5 + B_1 c_5 + C_1 = 0,  \\
            & A_2 s_4 + B_2 c_4 + C_2 = 0,  \\
            & (p_y^2 + p_z^2)s_1 - A_3 p_y - B_3 p_z = 0, 
        \end{cases}
        \label{eq:ik_q541}
    \end{equation}
    where
    \begin{equation}
        \begin{cases}
            & A_1 = 2 l_1 (l_2 + l_4) s_{\beta}, \\
            & B_1 = 2 l_3 (l_2 + l_4), \\
            & C_1 = l_1^2 + (l_2 + l_4)^2 + l_3^2 - (p_x^2 + p_y^2 + p_z^2), \\
            & A_2 = l_3 + (l_2 + l_4) c_5,  \\
            & B_2 = (l_2 + l_4) c_{\beta} s_5, \\
            & C_2 = p_x, \\
            & A_3 = -(l_2 + l_4) c_{\beta} s_4 s_5 + (l_2 + l_4) c_4 c_5 + l_3 c_4,  \\
            & B_3 = l_1 + (l_2 + l_4) s_{\beta} s_5.
        \end{cases} 
    \end{equation}
    Finally, the inverse kinematics of the 3-DoF spatial serial RRR manipulator are as follows
    \begin{equation}
        \begin{cases}
            & q_5 = \atantwo(-C_1, \pm \sqrt{A_1^2 + B_1^2 - C_1^2}) - \atantwo(B_1, A_1),\\
            & q_4 = \atantwo(-C_2, \pm \sqrt{A_2^2 + B_2^2 - C_2^2}) - \atantwo(B_2, A_2),\\
            & q_1 = \arcsin{(\frac{p_y A_3 + p_z B_3}{p_y^2 + p_z^2})}.
        \end{cases}
    \end{equation}
    Note that the above equations theoretically have four sets of solutions. We should choose a feasible solution according to the physical constraints of linkage.

    As we have obtained the solutions of $q_1$, $q_4$, and $q_5$, the active joint angles can be calculated by the following equation
    \begin{equation}
        q_a = \begin{bmatrix}
            q_1 \\
            q_4 + 2 \atan(k\cot{(\frac{q_5}{2})}) + \pi \\
            q_4
        \end{bmatrix}.
        \label{eq:q2qa}
    \end{equation}
    
\subsection*{Differential Kinematics}

    Similar to the displacement mapping, the velocity mapping from the active joint space to the foot space can also be obtained in two steps:
    \begin{enumerate}
        \item Find the mapping between the spanning tree's active and passive joint velocities.
        \item Find the mapping between $(\Dot{q}_1, \Dot{q}_4, \Dot{q}_5)$ and foot velocity.
    \end{enumerate}
    The angular velocity of the joints of the spanning tree model can be mapped from the active joints by
    \begin{equation}
        \Dot{q} = G(q_a) \Dot{q}_a,
        \label{eq:dqa2dq} 
    \end{equation}
    where $G(q_a)$ is the analytical Jacobian describing the velocity mapping between $q_a$ and $q$, and can be obtained directly by differentiating Eq. \eqref{eq:qa2q}
    \begin{align}
        G(q_a) & = \begin{bmatrix}
            1 & 0 & 0 \\
            0 & 1 & 0 \\
            0 & -h & h \\
            0 & 0 & 1 \\
            0 & -h & h
        \end{bmatrix},  
        \\
        h & = \frac{k(1 + t_{\eta}^2)}{(1 + k^2 t_{\eta}^2)}. 
    \end{align}

    The analytical Jacobian of 3-DoF spatial serial RRR manipulator can be derived by geometric Jacobian
    \begin{equation}
        J_{serial}(q_1, q_4, q_5) = \begin{bmatrix}
            -[p] & I_{3 \times 3}
        \end{bmatrix} \begin{bmatrix}
            J_{s1} & J_{s2} & J_{s3}
        \end{bmatrix},
    \end{equation}
    where $p$ is the foot position and can be calculated by Eqs. \eqref{eq:qa2q} and \eqref{eq:PoE}, $[p]$ is the skew symmetric matrix of $p$, $I_{3 \times 3}$ is the identity matrix, $J_{si}$ represents the screw of the $i$th joint, and satisfies $J_{s1} = \mathcal{S}_1$, $J_{s2} = Ad_{e^{[\mathcal{S}_1]q_1}}(\mathcal{S}_2)$, and $J_{s3} = Ad_{e^{[\mathcal{S}_1]q_1 \mathcal{S}_2]q_4}}(\mathcal{S}_3)$. $Ad_T$ represents the adjoint matrix based on the homogeneous matrix $T$.

    Since the velocity mappings are linear, the analytical Jacobian from the angular velocity of active joints to the foot velocity can be written as
    \begin{equation}
        J_{Bennett}(q_1, q_2, q_4) = J_{serial} \begin{bmatrix}
            1 & 0 & 0 \\
            0 & 0 & 1 \\
            0 & -h & h
        \end{bmatrix},
    \end{equation}
    where $J_{serial} \in \mathbb{R}^{3 \times 3}$ is the analytical Jacobian of the 3-DoF spatial serial RRR manipulator.

\subsection*{Inverse Dynamics}

    The dynamics of closed-loop mechanisms are different from those of open-chain mechanisms, and the additional effects of the closed joint need to be considered. The inverse dynamics computation is divided into three steps:
    \begin{enumerate}
        \item Calculate the spanning tree model's joint angles, angular velocities, and accelerations.
        \item Solve the inverse dynamics of the spanning tree model.
        \item Map the joint torques of the spanning tree model to the active joint torques by considering the closed joint.
    \end{enumerate}

    The spanning tree model's joint angles and angular velocities can be calculated by Eqs. \eqref{eq:qa2q} and \eqref{eq:dqa2dq}. The angular accelerations are computed as follows
    \begin{equation}
        \Ddot{q} = \Dot{G}(q_a) \Dot{q}_a + G(q_a) \Ddot{q}_a,
        \label{eq:ddqa2ddq}
    \end{equation}
    where 
    \begin{align}
        \Dot{G}(q_a, \Dot{q}_a) & = \begin{bmatrix}
            0 & 0 & 0 \\
            0 & 0 & 0 \\
            0 & -\Dot{h} & \Dot{h} \\
            0 & 0 & 0 \\
            0 & -\Dot{h} & \Dot{h}
        \end{bmatrix},  \\
        \Dot{h} & = \frac{k(k^2 - 1) t_{\eta}}{(1 + k^2 t^2_{\eta})^2 c^2_{\eta}} (\Dot{q}_2 - \Dot{q}_4). 
    \end{align}

    The inverse dynamics of the spanning tree model can be written as 
    \begin{equation}
        \tau = M(q) \Ddot{q} + C(q,\Dot{q}) \Dot{q} + g(q),
        \label{eq:ID_spanning_tree}
    \end{equation}
    where $M(q) \in \mathbb{R}^{5 \times 5}$ is the inertia matrix, $C(q,\Dot{q}) \in \mathbb{R}^{5 \times 5}$ is the Coriolis and centrifugal matrix, and $g(q) \in \mathbb{R}^{5 \times 1}$ is the gravitational terms. 

    Force Jacobian can compute the active joint torques between the active joints and the joints of the spanning tree model~\cite{ghorbel2000modeling}
    \begin{equation}
        \tau_a = G^T(q_a) \tau, 
        \label{eq:tau2taua}
    \end{equation}
    where $\tau_a$ denotes the active joint torque and $\tau$ denotes the joint torque of spanning tree model. 
    
    By combining Eqs. \eqref{eq:qa2q}, \eqref{eq:dqa2dq}, \eqref{eq:ddqa2ddq}, \eqref{eq:ID_spanning_tree}, and \eqref{eq:tau2taua}, the inverse dynamics of ORL can be written as 
    \begin{equation}
        \tau_a = M_a(q_a) \Ddot{q}_a + C_a(q_a, \Dot{q}_a) \Dot{q}_a + g_a[\gamma(q_a)], \label{eq:ID_bennett}
    \end{equation}
    where $ M_a(q_a) \in \mathbb{R}^{3 \times 3}$ is the inertia matrix of active joints, $ C_a(q_a, \Dot{q}_a) \in \mathbb{R}^{3 \times 3}$ is the Coriolis and centrifugal matrix of active joints, and $g_a(\gamma(q_a)) \in \mathbb{R}^{3 \times 1}$ is the gravitational terms of active joints, satisfying
    \begin{equation}
        \begin{cases}
            & M_a(q_a) = G^T(q_a) M[\gamma(q_a)] G(q_a),  \\
            & C_a(q_a, \Dot{q}_a) = G^T(q_a) M[\gamma(q_a)] \Dot{G}(q_a) \\
                                 & \quad \quad \quad \quad \quad + G^T(q_a) C[\gamma(q_a),G(q_a) \Dot{q}_a] G(q_a),\\
            & g_a[\gamma(q_a)] = G^T(q_a) g[\gamma(q_a)].
        \end{cases}
    \end{equation}

\vspace{0.5cm}
\section*{Appendix C on Overconstrained Quadruped Control}
\label{App-C:Controller}

    The state estimation mainly obtains the position and velocity of the robot's CoM, while the rotation and angular velocity can be obtained through the IMU. Based on the contact invariant, we estimate the CoM state by fusing IMU and leg kinematics via the Kalman filter~\cite{Flayols2017Experimental}. We also implement the terrain estimation, enabling the overconstrained quadruped to adjust the posture and move on the slope or stairs~\cite{Gehring2015Dynamic}.
    
    The motion generation module would output the desired trajectory of the center of mass (CoM) to the stance phase controller, output the desired foot trajectory to the swing phase controller, and output the contact state to determine the phase of each leg. The CoM planner generally gives the high-level trajectory, considering the efficiency~\cite{Chen2020Optimized} or stability~\cite{Kalakrishnan2010Fast}. Focusing on the robot's locomotion performance, the user joystick's velocity commands are used, and the position commands are integrated from velocity. The contact state of the foot is generated by a fixed-timing-based gait scheduler that reduces the system complexity brought by contact sensors. The leg runs a stance phase controller when a foot is scheduled to contact. Otherwise, the leg would liftoff and run a swing phase controller. 

    The model predictive controller is chosen as the stance phase controller, widely used and verified to be robust and suitable for various tasks~\cite{Di2018Dynamic, Ding2021Representation, Neunert2018Whole}. The swing phase control requires foothold selection, foot trajectory planning, and foot trajectory tracking.

\subsection*{The Simplified Dynamics Model}

\begin{figure}[!b]
        \centering
        \includegraphics[width=1\columnwidth]{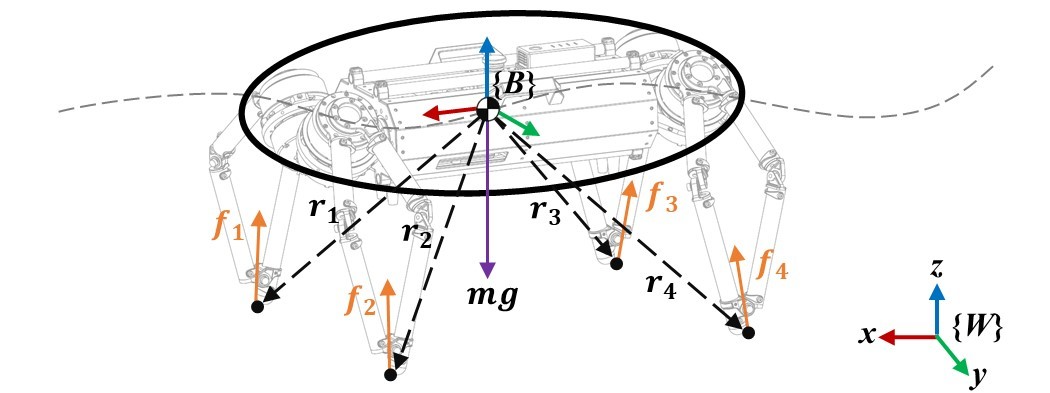}
        \caption{
        Schematic diagram of the single rigid body model.
        }
        \label{fig:SRBD}
    \end{figure}

    Due to the low-inertia leg design, the overconstrained quadruped can be modeled as a single rigid body subject to the Ground Reaction Force (GRF), as shown in Fig.~\ref{fig:SRBD}. The spatial force on a single rigid body is equal to the differential of the spatial momentum, and the single rigid body dynamics model is formulated as
    \begin{align}
        \mathcal{F} & = \frac{d}{dt} (\mathcal{IV}) = \mathcal{IA} + \mathcal{V} \times^{*} \mathcal{IV} \nonumber \\
        & = \begin{bmatrix} 
        I \Dot{\omega} + \omega \times  (I\omega) \\ m \Ddot{p}_{com} 
        \end{bmatrix}, \label{eq:SRBD} 
    \end{align}
    where $\mathcal{F} \in \mathbb{R}^{6 \times 1}$ is the spatial wrench, $\mathcal{I} \in \mathbb{R}^{6 \times 6}$ is the spatial inertia, $I \in \mathbb{R}^{3 \times 3}$ is the inertia tensor, $m \in \mathbb{R}$ is the mass, $\mathcal{V} \in \mathbb{R}^{6 \times 1}$ is the spatial velocity, $\mathcal{A} \in \mathbb{R}^{6 \times 1}$ is the spatial acceleration, $\omega \in \mathbb{R}^{3 \times 1}$ and $\Dot{\omega} \in \mathbb{R}^{3 \times 1}$ are the angular velocity and angular acceleration, and $\Ddot{p}_{com} \in \mathbb{R}^{3 \times 1}$ is the velocity of CoM. The $\omega \times  (I\omega)$ of the Euler term can be ignored when the angular velocity is small, and the Euler term of Eq. \eqref{eq:SRBD} can be approximated with $I \Dot{\omega}$. The simplified dynamics model of the overconstrained quadruped in world frame is given by
    \begin{equation}
        \begin{bmatrix}
            \sum_{j=1}^{4} (r_j \times f_j) \\
            mg + \sum_{j=1}^{4} f_j 
        \end{bmatrix} = 
        \begin{bmatrix} 
        I  \Dot{\omega} \\ 
        m  \Ddot{p}_{com} 
        \end{bmatrix},
        \label{eq:SRBD_quadruped}
    \end{equation}
    where $f_j \in \mathbb{R}^{3 \times 1}$ is the GRF of the $j_{th}$ leg, and $r_j \in \mathbb{R}^{3 \times 1}$ is the vector from CoM to the contact point of $f_j$.

    The state of robot consists of the Euler angle $\Theta \in \mathbb{R}^{3 \times 1}$, position of CoM $p_{com} \in \mathbb{R}^{3 \times 1}$, angular velocity $\omega \in \mathbb{R}^{3 \times 1}$, velocity of CoM $\Dot{p}_{com} \in \mathbb{R}^{3 \times 1}$, and the third term of gravity $g(3) \in \mathbb{R}$. The control input of the model is the GRF of four legs, and the state and control input of the model is defined as 
    \begin{align}
        x & = \begin{bmatrix}
            \Theta^T & p_{com}^T & \omega^T & \Dot{p}_{com}^T & g(3)
        \end{bmatrix}^T, \label{eq:state} \\
        u & = \begin{bmatrix}
            f_1^T & f_2^T & f_3^T & f_4^T
        \end{bmatrix}^T. \label{eq:input}
    \end{align}
    The simplified single rigid body dynamics can be formulated as a discrete-time state-space model, the state matrix $A_k \in \mathbb{R}^{13 \times 13}$ and the input matrix $B_k \in \mathbb{R}^{13 \times 12}$ is defined as
    \begin{align}
        A_k & = 
        \begin{bmatrix}
            \mathbb{I}_3 & 0_{3 \times 3} & R_z^T(\psi_{k}) \Delta T & 0_{3 \times 3} & 0_{3 \times 1} \\
            0_{3 \times 3} & \mathbb{I}_3 & 0_{3 \times 3} & \mathbb{I}_3 \Delta T & 0_{3 \times 1} \\
            0_{3 \times 3} & 0_{3 \times 3} & \mathbb{I}_3 & 0_{3 \times 3} & 0_{3 \times 1} \\
            0_{3 \times 3} & 0_{3 \times 3} & 0_{3 \times 3} & \mathbb{I}_3 & \begin{bmatrix} 0 \\ 0 \\ \Delta T \end{bmatrix} \\
            0_{1 \times 3} & 0_{1 \times 3} & 0_{1 \times 3} & 0_{1 \times 3} & 1
        \end{bmatrix}, \\
        B_k & =
        \begin{bmatrix}
            0_{3 \times 3} & 0_{3 \times 3} & 0_{3 \times 3} & 0_{3 \times 3} \\
            0_{3 \times 3} & 0_{3 \times 3} & 0_{3 \times 3} & 0_{3 \times 3} \\
            I^{-1}_k [r_1] & I^{-1}_k [r_2] & I^{-1}_k [r_3] & I^{-1}_k [r_4] \\
            \mathbb{I}_3 \Delta T/m & \mathbb{I}_3 \Delta T/m & \mathbb{I}_3 \Delta T/m & \mathbb{I}_3 \Delta T/m \\
            0_{1 \times 3} & 0_{1 \times 3} & 0_{1 \times 3} & 0_{1 \times 3}  
        \end{bmatrix},
    \end{align}
    where $\Delta T$ is the control period, $R_z(\psi_{k}) \in \mathbb{R}^{3 \times 3}$ is the rotation matrix depend on yaw at $k_{th}$ step, $x_k \in \mathbb{R}^{13 \times 1}$ is the state at $k_{th}$ step, $u_k \in \mathbb{R}^{12 \times 1}$ is the control input at $k_{th}$ step, $\mathbb{I}_3 \in \mathbb{R}^{3 \times 3}$ is the identity matrix, and $0_{i \times j} \in \mathbb{R}^{i \times j}$ is the zero matrix.

\subsection*{Further Results on Simulation with MPC}

\begin{figure}[htbp]
        \centering
        \includegraphics[width=\linewidth]{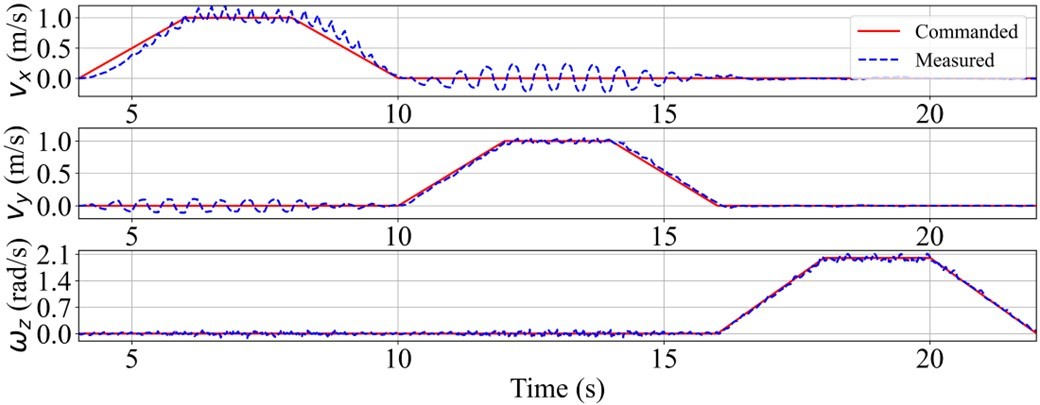}
        \caption{
        \textbf{The velocities of the overconstrained quadruped when omnidirectional trotting.}
        The red and blue solid lines are the commanded and measured values.
        }
        \label{fig:omni_vel}
    \end{figure}

    To obtain a smooth velocity, the commanded trajectories are uniformly accelerated to a specified velocity, then keep a constant velocity, and finally uniformly decelerated to zero. The 1 m/s forward velocity, 1 m/s lateral velocity, and 2 rad/s angular velocity are achieved in simulation, as shown in Fig.~\ref{fig:omni_vel}. 
    
\begin{figure}[htbp]
        \centering
        \includegraphics[width=\linewidth]{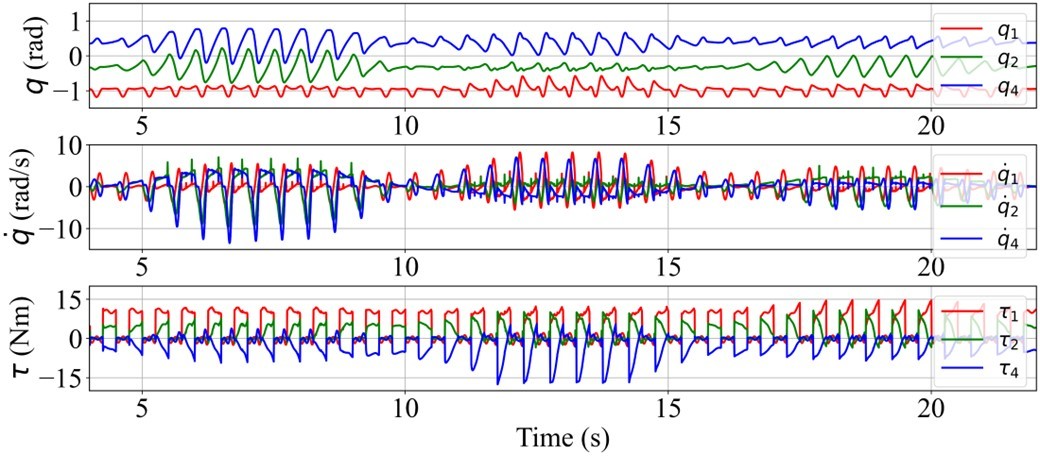}
        \caption{
        \textbf{The joint position (top), velocity (middle), and torque (bottom) of the front right leg when omnidirectional trotting.}
        }
        \label{fig:omni_joint}
    \end{figure}

\begin{figure}[htbp]
        \centering
        \includegraphics[width=\linewidth]{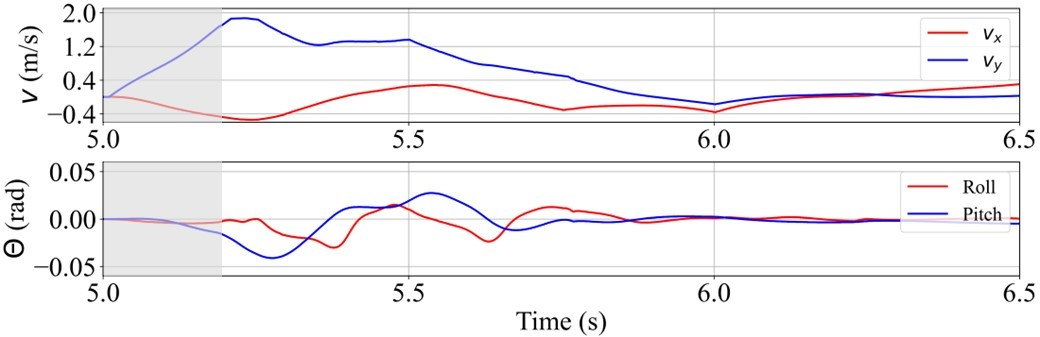}
        \caption{
        \textbf{The velocity (top) and Euler angle (bottom) of the overconstrained quadruped when push recovery.}
        The gray part indicates the time when the external force is applied.
        }
        \label{fig:push}
    \end{figure}

\begin{figure}[htbp]
        \centering
        \includegraphics[width=\linewidth]{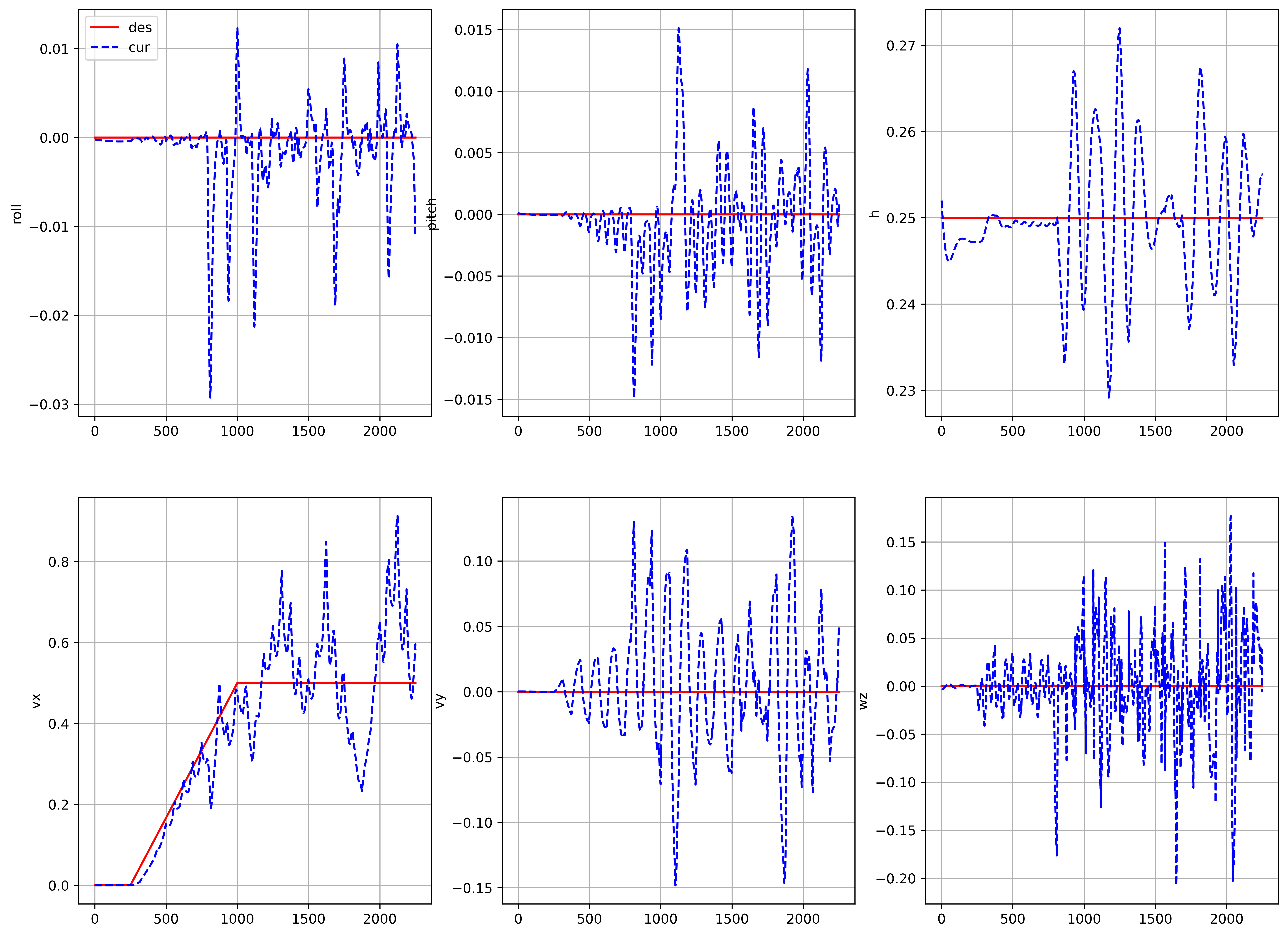}
        \caption{
        \textbf{The state of the overconstrained robot compared with the desired input states trotting on rugged terrain.}
        }
        \label{fig:rugged_terrain}
    \end{figure}

\begin{figure}[htbp]
        \centering
        \includegraphics[width=\linewidth]{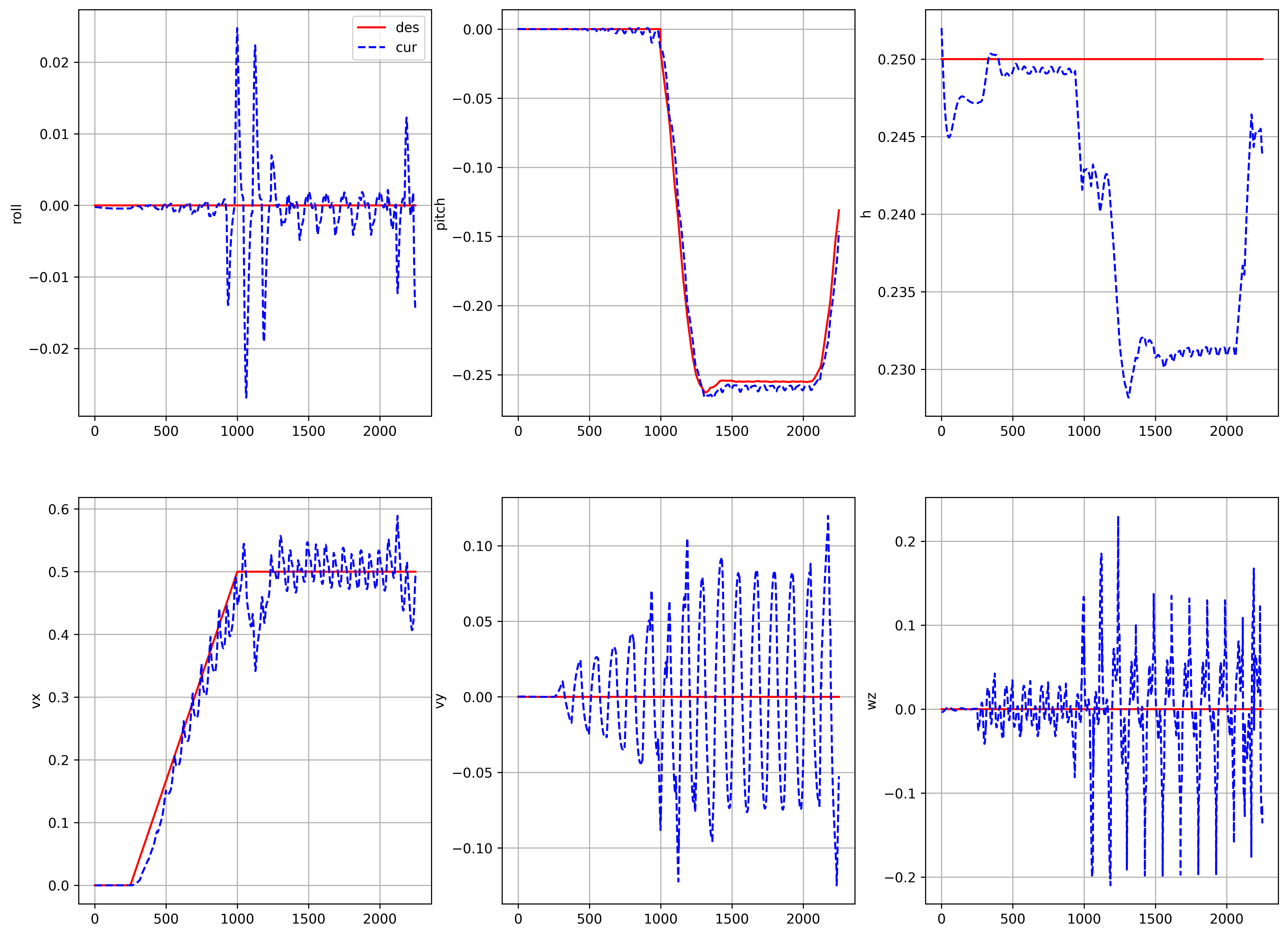}
        \caption{
        \textbf{The state of the overconstrained robot compared with the desired input states trotting on sloped terrain.}
        }
        \label{fig:slope}
    \end{figure}

\begin{figure}[htbp]
        \centering
        \includegraphics[width=\linewidth]{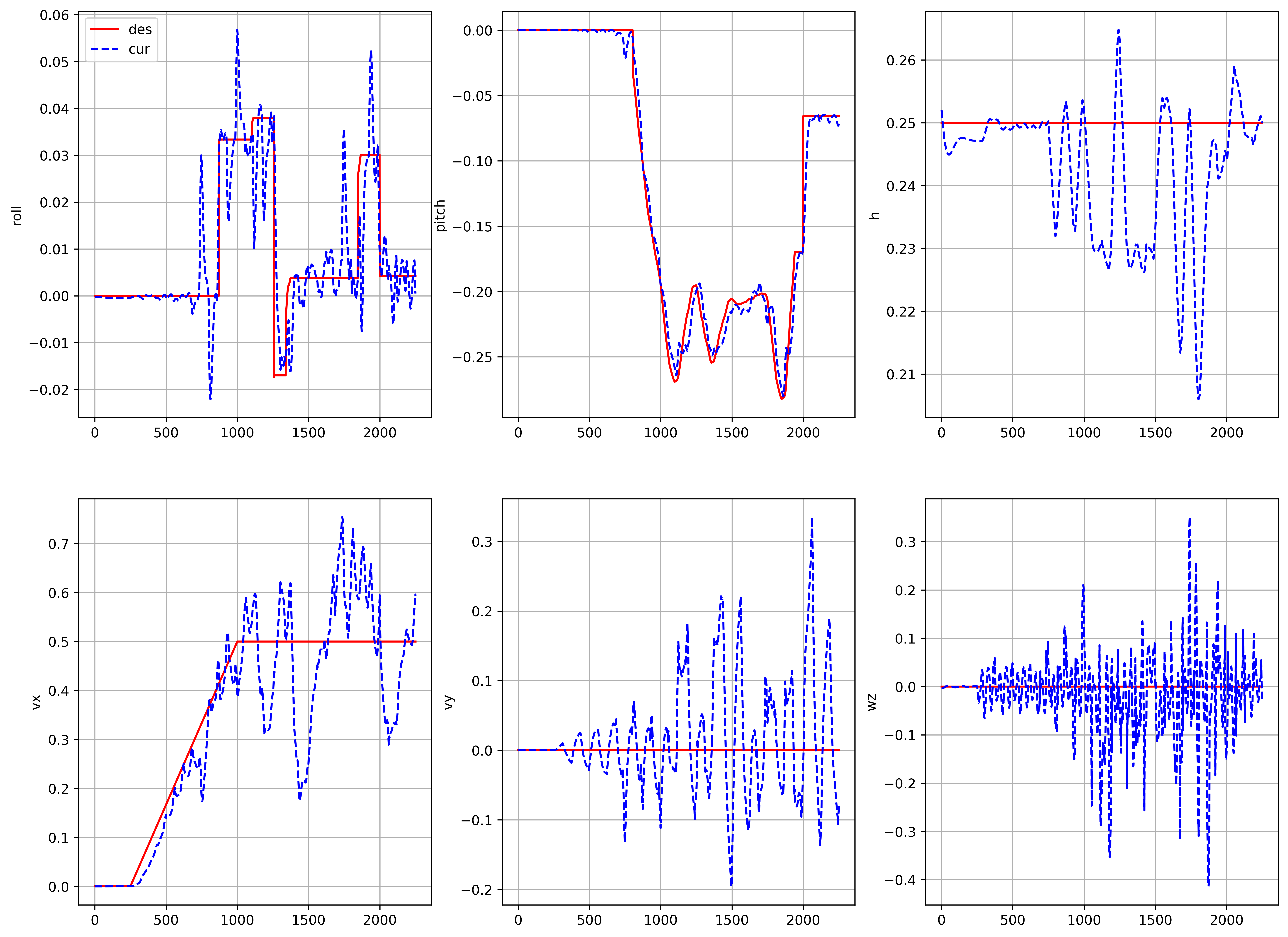}
        \caption{
        \textbf{The state of the overconstrained robot compared with the desired input states trotting on stairs.}
        }
        \label{fig:stairs}
    \end{figure}

    Fig.~\ref{fig:omni_joint} shows the front right leg's joint position, velocity, and torque during trotting. The peak joint velocity of forward trotting is more significant, while the greater peak joint torque occurs during lateral trotting. The peak joint velocity and torque are below the actual motor limit, and it can potentially increase the walking speed. The changes in the velocity and Euler angles are shown in Fig.~\ref{fig:push}. Fig.~\ref{fig:rugged_terrain}, Fig.~\ref{fig:slope}, and Fig.~\ref{fig:stairs} show the roll, pitch, height, xy velocity, and yaw angular velocity of the overconstrained robot trotting on various terrains.

\section*{Appendix D on Reinforcement Learning}
\label{App-D:Learning}

    For flat terrain locomotion, we train the robot to track the command velocity along a single direction (forward or lateral) in world coordinates so that the robot can trot along a straight line, making it convenient for us to calculate COT. The reward functions and scales used are listed above. We use a relatively high joint penalty weight as energy efficiency is the major metric besides velocity tracking performance, and we lower other penalty terms, such as action rate penalty, angular velocity, and z-direction velocity penalty. We keep a high foot air time reward weight, which we found crucial to generate a smooth gait pattern and mitigate robot feet scuffling in flat-terrain settings. We also found that a reasonably high-foot air time reward could prevent the robot from stocking at a local optimum, considering energy cost. Multi-terrain locomotion aims to track the command velocity of the robot base while traversing various challenging terrains. The reward functions and scales are listed below. 

\subsection*{Reward Functions for Flat-terrain Locomotion}

    $x$ velocity tracking (forward only): 
    \begin{itemize}
        \item $\exp\{-|^{w}v_{x} -v_{x}^{cmd}|/0.25\} * 1.0$
    \end{itemize}

    $y$ velocity tracking (lateral only):
    \begin{itemize}
        \item $\exp\{-|^{w}v_{y} -v_{y}^{cmd}|/0.25\} * 1.0$
    \end{itemize}

    dof energy:
    \begin{itemize}
        \item $\sum_{j=1}^{12} \Delta\theta_{j}\tau_{j} * -0.2$
    \end{itemize}

    action rate:
    \begin{itemize}
        \item $|a_{t-1}-a_t|^2 * -0.0025$
    \end{itemize}

    feet air time:
    \begin{itemize}
        \item $\sum{(t_{air}-0.5)*\mathbbm{1}_{new\_contact}} * 1.5$
    \end{itemize}

    orientation:
    \begin{itemize}
        \item $|\Delta{\theta}| * -0.5$
    \end{itemize}

    angular velocity:
    \begin{itemize}
        \item $|^{w}\omega|^2 * -0.015$
    \end{itemize}

    $z$ velocity:
    \begin{itemize}
        \item $|^w{v_{z}}|^2 * -2.0$
    \end{itemize}

    $x$ distance (lateral only):
    \begin{itemize}
        \item $|x| * -0.1$
    \end{itemize}

    $y$ distance (forward only):
    \begin{itemize}
        \item $|y| * -0.1$
    \end{itemize}

\subsection*{Reward Functions for Multi-terrain Locomotion}

    $xy$ velocity tracking (forward only):
    \begin{itemize}
        \item $\exp\{-|^{b}v_{xy} -v_{xy}^{cmd}|/0.25\} * 1.5$
    \end{itemize}

    yaw velocity tracking:
    \begin{itemize}
        \item $\exp\{-|^{b}\omega_{z} -\omega_{z}^{cmd}|/0.25\} * 0.75$
    \end{itemize}

    dof energy:
    \begin{itemize}
        \item $\sum_{j=1}^{12} \Delta\theta_{j}\tau_{j} * -0.1$
    \end{itemize}

    action rate:
    \begin{itemize}
        \item $|a_{t-1}-a_t|^2 * -0.01$
    \end{itemize}

    feet air time:
    \begin{itemize}
        \item $\sum{(t_{air}-0.5)*\mathbbm{1}_{new\_contact}} * 0.01$
    \end{itemize}

    roll-pitch velocity:
    \begin{itemize}
        \item $|^{b}\omega_{xy}|^2 * -0.05$
    \end{itemize}

    $z$ velocity:
    \begin{itemize}
        \item $|^w{v_{z}}|^2 * -2.0$
    \end{itemize}

\subsection*{Grid Search Hyperparameter for Flat-Terrain Locomotion}

    The performance of the quadruped robot may depend on the hyperparameter selection. To make the comparison fair enough, we perform a grid search on the parameters listed below and train each parameter combination for 1,000 policy updates.

    \begin{itemize}
        \item timesteps per rollout: $24, 48$
        \item epochs per rollout: $5, 7$
        \item minibatches per epoch: $2, 4$
        \item actor/critic hidden layers: $[64, 64, 64], [128, 128, 128]$
    \end{itemize}

\subsection*{Training Parameters for Multi-Terrain Locomotion}

    We kept the same parameters for multi-terrain locomotion training listed below. For multi-terrain locomotion, we trained each robot type for 2,000 policy updates.

    \begin{itemize}
        \item discount factor: $0.99$
        \item timesteps per rollout: $48$
        \item epochs per rollout: $7$
        \item minibatches per epoch: $4$
        \item entropy bonus: $0.01$
        \item value loss coefficient: $1.0$
        \item clip range: $0.2$
        \item learning rate: $1e-3$
        \item environment number: $4096$
        \item initial noise standard deviation: $1.0$
        \item learning rate schedule: \texttt{Adaptive}
        \item desired KL divergence: $0.01$
        \item maximum gradient norm: $1.0$
        \item actor network hidden layers: $[128, 128, 128]$
        \item critic network hidden layers: $[128, 128, 128]$
        \item activation function: \texttt{ELU}
    \end{itemize}

\end{document}